%% file: acl_latex.tex
\pdfoutput=1

\documentclass[11pt]{article}

\usepackage[preprint]{acl}

\usepackage{times}
\usepackage{latexsym}
\usepackage{listings}
\usepackage{hyperref}

\usepackage[T1]{fontenc}

\usepackage[utf8]{inputenc}

\usepackage{microtype}

\usepackage{inconsolata}
\usepackage{multirow}
\usepackage{arydshln}
\usepackage{amsmath}
\usepackage{booktabs}
\usepackage{xcolor} 

\usepackage{graphicx}

\colorlet{punct}{red!60!black}
\definecolor{background}{HTML}{EEEEEE}
\definecolor{delim}{RGB}{20,105,176}
\colorlet{numb}{magenta!60!black}
\definecolor{comment}{HTML}{999999}
\definecolor{charname}{HTML}{0070C0} 

\lstdefinelanguage{json}{
    basicstyle=\normalfont\ttfamily\small,
    numberstyle=\scriptsize,
    showstringspaces=false,
    breaklines=true,
    frame=lines,
    backgroundcolor=\color{background},
    literate=
     *{0}{{{\color{numb}0}}}{1}
      {1}{{{\color{numb}1}}}{1}
      {2}{{{\color{numb}2}}}{1}
      {3}{{{\color{numb}3}}}{1}
      {4}{{{\color{numb}4}}}{1}
      {5}{{{\color{numb}5}}}{1}
      {6}{{{\color{numb}6}}}{1}
      {7}{{{\color{numb}7}}}{1}
      {8}{{{\color{numb}8}}}{1}
      {9}{{{\color{numb}9}}}{1}
      {:}{{{\color{punct}{:}}}}{1}
      {,}{{{\color{punct}{,}}}}{1}
      {\{}{{{\color{delim}{\{}}}}{1}
      {\}}{{{\color{delim}{\}}}}}{1}
      {[}{{{\color{delim}{[}}}}{1}
      {]}{{{\color{delim}{]}}}}{1},
}

\definecolor{myTitleBackground}{HTML}{F0F0F0}
\definecolor{mySeparatorLine}{HTML}{C0C0C0} 

\usepackage{enumitem}
\usepackage[most]{tcolorbox}
\newcommand{\dialoguefont}{\normalfont\rmfamily\small} 

\tcbset{
    dialoguecommon/.style={ 
        enhanced,
        boxrule=0.5pt,
        arc=4pt,
        outer arc=4pt,
        drop shadow={opacity=0.3, fill=black!50!white},
        boxsep=5pt,
        left=7pt,right=7pt,top=10pt,bottom=5pt
    }
}

\newtcolorbox{dialogueboxA}[1]{
    dialoguecommon, 
    colback=blue!2!white,      
    colframe=blue!30!white,    
}

\newtcolorbox{dialogueboxB}[1]{%
    dialoguecommon, 
    colback=gray!10!white,    
    colframe=gray!40!white,  
}

\newtcolorbox{dialogueboxC}[1]{%
    dialoguecommon, 
    colback=green!6!white,     
    colframe=green!30!white,   
}

\newtcolorbox{dialogueboxD}[1]{%
    dialoguecommon, 
    colback=red!5!white,    
    colframe=red!30!white,  
}

\lstdefinelanguage{dialoguecsv}{
    basicstyle=\normalfont\ttfamily\small,
    numbers=none,
    breaklines=true,
    frame=lines,
    backgroundcolor=\color{background},
    columns=fullflexible, 
    keepspaces=true, 
    tabsize=4,
    showtabs=false,
    keywordstyle=\color{charname},
    keywords={Michael, Kenneth}, 
    upquote=true,
}

\newlist{dialoguelist}{itemize}{1}
\setlist[dialoguelist,1]{%
    label={},
    leftmargin=0.2em,
    labelsep=0pt,
    align=left,
    itemsep=0.5em,
    parsep=0pt,
    font=\dialoguefont,
    before=\setlength{\parskip}{0pt},
    after=\setlength{\parskip}{0pt}
}

\newcommand{\indiref}{\texttt{IndiRef}}

\title{Frame of Reference: Addressing the Challenges of Common Ground Representation in Situational Dialogs}

\author{Biswesh Mohapatra\thanks{Corresponding Authors} \\ Inria            \And
        Théo Charlot\thanks{\ \ Work done during internship at Inria.}\thanks{\ \ Authors contributed equally} \\ Nantes Université \And
        Giovanni Duca\footnotemark[2]\footnotemark[3] \\ University of Trento
        \AND
        Mayank Palan\footnotemark[2]\footnotemark[3] \\ VJTI Mumbai \\ \And
        Laurent Romary\footnotemark[1] \\ Inria \\ \And
        Justine Cassell\footnotemark[1] \\ Inria, Carnegie Mellon University 
        \AND
        \textnormal{\texttt{\{biswesh.mohapatra, laurent.romary, justine.cassell\}@inria.fr}}
        }

\begin{document}
\maketitle
\begin{abstract}
Common ground plays a critical role in situated spoken dialogs, where interlocutors must establish and maintain shared references to entities, events, and relations to sustain coherent interaction in a shared space and over time. With the increasing presence of embodied conversational agents and social robots, the ability to correctly ground this kind of conversational content in order to refer back later also becomes important for dialog systems. Prior studies have demonstrated that LLMs are capable of performing certain grounding acts like acknowledgments. However, relatively little work has investigated their capacity to leverage the grounded information, like in complex scenarios involving space and time (e.g., "let's go to that café near the park we went to yesterday"). To that end, in this work, we evaluate a model's ability to establish common ground by utilizing these "relational references" in the dynamic and shared environments of situated dialogs. We then test multiple methods for representing common ground and further propose approaches to improve their performance by using reinforcement learning on our synthetically generated dialog data .
\end{abstract}

\input{sections/introduction}
\vspace{-1mm}
\input{sections/related_work}
\vspace{-1mm}
\input{sections/metric}

\input{sections/baseline}
\vspace{-10mm}
\input{sections/results}

\input{sections/improve}

\input{sections/conclusion}

\section*{Limitations} 

Certain limitations offer avenues for future research. First, the dataset used for our test cases, while selected to approximate spontaneous spoken dialog, has inherent constraints. The ideal dataset would comprise spontaneous, multi-session spoken dialogs. The logistical challenges in acquiring such data necessitated the use of two specialized dialog corpora, Meetup and Spot the Difference that, while effective in providing a range of grounding phenomena through a game-based framework, may not entirely capture the nuances of spontaneous multi-topic and multi-session human interaction. 
Although the corpora may lack the full breadth of really long conversation, the dataset used for our corpora, especially Meetup, provides a critical testing ground by featuring fragmented context i.e. speakers discuss and recall information exchanged in previously visited simulated environments (different rooms), effectively mimicking the transition between distinct contexts. A resource-constrained setting using such dialogs thus allows us to rigorously test how different grounding representations sustain performance under such fragmented context constraints which are also inherent to long-term, multi-session interactions.


Second, due to computational resource limitations, our empirical evaluation was restricted to the 8-billion parameter Llama 3.1 instruction-tuned model especially for the GRPO-based training. A central hypothesis in current NLP research is that increasing model scale can reduce issues like hallucination. While we conjecture that our findings would benefit from this effect, validating this hypothesis was beyond the scope of our  resources. A valuable direction for future work would be to replicate and extend our experiments using larger models. 

Lastly, while our test cases provide a solid foundation for testing the model's ability to use established common ground, they do not comprehensively test it's capacity to handle `assumptions' and `beliefs'. These deeper cognitive elements, while tilting towards theory of mind research, are still essential for robust conversational grounding and present a clear direction for future investigation.

\section*{Acknowledgments}
This work was granted access to the HPC resources of IDRIS under the allocation [AD011016333] made by GENCI.

\bibliography{custom}

\appendix
\label{sec:appendix}
\input{sections/appdx}

\end{document}

%% file: sections/introduction.tex
\section{Introduction}

In dialog, common ground refers to the shared knowledge, beliefs, and assumptions that accumulate between participants as a conversation progresses~\cite{Clark1989ContributingTD}. Conversational Grounding is the process of building and maintaining this shared understanding. While recent advances in Large Language Models (LLMs) have demonstrated proficiency in performing certain grounding acts~\cite{TraumGA} such as acknowledging inputs for immediate context~\cite{Mohapatra2024EvaluatingTE} - it remains unclear whether these behaviors represent genuine understanding or merely an ``illusion of grounding''. This distinction is critical: a system may mimic understanding through plausible acknowledgments, yet fail to utilize the information later effectively.

\input{figures/cg}

 This challenge is amplified in dialogs that extend over longer periods. As dialog history grows, systems must move beyond context windows and employ memory management techniques to retrieve information from the established common ground. In order to give the systems a true grounding ability, the acknowledged information must therefore be represented and stored in a format that can be utilized later on, regardless of the size of the dialog. Its importance is particularly pronounced in spoken situational dialogs~(dialogs that operate in the dynamic and shared environments) such as Embodied Conversational Agents (ECAs)~\cite{CASSELL200155} and Social Robots~\cite{soc_rob} where the conversations can take place over multiple interactions. To become effective collaborators, these agents must be able to deploy information from their ongoing and previous interactions.

This gives rise to three central research questions that we address in this paper : 
\begin{enumerate}
    \item \textbf{RQ1 Benchmarking:}  How do we measure the capability of a dialog system to establish a persistent, usable, and useful common ground?
    \item \textbf{RQ2 Representation:} In real-world scenarios where dialog history exceeds context window size, how well do commonly used common ground representations perform in this benchmark?
    \item \textbf{RQ3 Improving Performance:} Given the potential inadequacy of such representations, how can we improve our system's grounding capabilities?
\end{enumerate}

\vspace{2mm}
\noindent \textbf{Key Contributions and Findings:}

\begin{itemize}
    \item \textbf{Benchmarking Grounding Capabilities:} \citet{kruijt-vossen-2022-role} demonstrated that in situated dialogs, people often refer to established common ground using references containing relations to specific properties, entities, or events. In this paper, we group such references under the rubric of `relational references'. To address RQ1, we thus propose that a system's ability to resolve such complex relational references established through prior dialog can serve as a measure of its capability to create persistent, useful, and usable common ground.

For instance, in Figure~\ref{fig:cg1}, the user refers to the phone via the shop where it was bought. Such recall of entities often lack a unique reference and could equally be described temporally (e.g., first shop) instead of spatially (e.g., shop next to the church). 
Without a grounding capability that establishes a persistent and informational common ground~(whether internal or external) that can be utilized later, a dialog system cannot successfully resolve these relational references.

We validate \citet{kruijt-vossen-2022-role} by examining two natural dialog corpora to identify how humans leverage common ground, finding multiple examples of such complex relational references. Informed by this, we introduce a new evaluation benchmark, \indiref. It moves beyond the immediate simple grounding acts such as acknowledgments and creates a benchmark to evaluate whether dialog systems can resolve complex references to the established common ground (e.g., ``that restaurant next to the museum that we went to in New York''), a critical requirement for effective long-term conversation in spoken situated dialog. 
    
    \item \textbf{Evaluation of Current Techniques:} To answer RQ2, we evaluate LLMs of varying sizes in both full-dialog context~(as a baseline) and resource-constrained settings (mimicking real-life longer conversations exceeding context lengths of LLMs). Our experiments reveal that standard LLMs struggle with relational references in spoken situated dialogs, even with full context. We also test the performance of multiple commonly used common ground representation techniques for resource-constrained settings. However, they often perform worse than LLMs with full-context due to their failure to capture the relationships between entities which are necessary for accurate retrieval.
    
    \item \textbf{Synthetic Data \& RL Training:} To address RQ3, we identify a lack of multi-step reasoning capability for spoken dialogs in current LLMs along with a scarcity of high-quality training data. We develop a synthetic data generation pipeline and show that training models on these data using Reinforcement Learning (RL) improves their ability to handle complex references.
\end{itemize}

By providing a benchmark to test the effectiveness of the common ground established by the dialog system using relational references, exposing the limitations of current techniques, and demonstrating the efficacy of RL-based training using synthetic data, this work provides a pathway toward dialog systems capable of robust, situated conversational grounding over time.

%% file: figures/cg.tex
\begin{figure}[t]
  \centering
  \includegraphics[width=0.8\columnwidth]{
  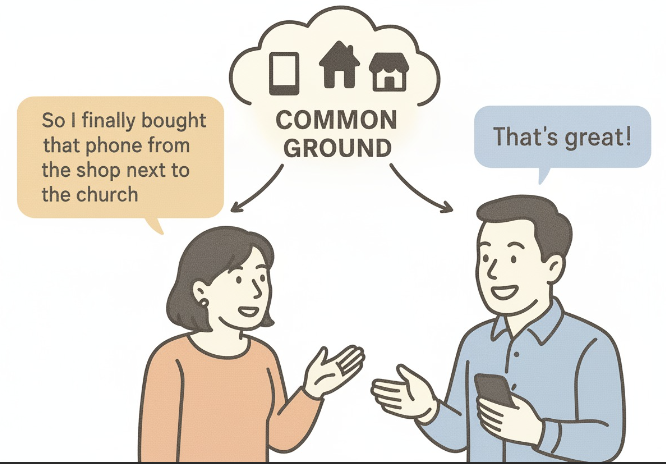}
\caption{An example of people referring back to the common ground.}
  \label{fig:cg1}
\end{figure}

%% file: sections/related_work.tex
\section{Related Work}

\citet{TraumGA} introduced the concept of Grounding Acts (GAs) that represent fundamental conversational moves like acknowledging or repairing.  \citet{Mohapatra2024EvaluatingTE} showed that modern LLMs can successfully perform some of these GAs when the required context is limited to the last few utterances. The practical challenges of connecting these local grounding decisions to the long-term maintenance of a coherent common ground remain significant \cite{Mohapatra2024ConversationalGA}.

Earlier attempts to represent and manage the common ground, such as Domain Reference theory~\cite{DRT}, were limited in nature due to their reliance on rule-based approaches. Foundational work has largely focused on structured, symbolic approaches for task-oriented dialogs. Methods like Dialog State Tracking (DST)~\cite{jacqmin-etal-2022-follow} and frame-based systems~\cite{frames} represent conversational context as slots and values, capturing the user's goals and constraints. While highly interpretable and effective for well-defined tasks, these approaches often lack the flexibility to handle the dynamic and unpredictable nature of long-term natural interactions.

More flexible representations such as dynamic knowledge graphs~(KGs)~\cite{graphWOZ, bu-etal-2025-query} have been proposed to model entities and their relationships as they emerge in conversation. KGs allow for a richer, more interconnected model of context and can naturally handle entity-based references by linking mentions to nodes in the graph. However, a significant challenge arises in spoken dialogs where entities often lack a single, stable referring expression (e.g. `park next to the restaurant' or `duck park') or relation (e.g., `the room that has a TV' and `TV is next to a table' implies `room has a table'). 

Neural memory architectures~\cite{Henaff2016TrackingTW, rajendran-etal-2019-learning, wu-yu-2024-stateful} encode dialog history in distributed vector representations. However, this "black-box" nature, makes it difficult to verify whether their conversational behaviors, such as acknowledgments, stem from a truly grounded understanding or pattern matching, especially, when there is an absence of an evaluation metric. 

Beyond tracking facts, a robust common ground representation requires modeling participant beliefs and perspectives~\cite{Traum2003, schlangen-skantze-2009-general}, necessitating a shift from state tracking to belief tracking. Further frameworks~\cite{butter, khebour-etal-2024-common, minddial} have proposed to improve reliability by verifying mutual understanding through the tracking of shared facts, evidence, and unresolved information. However, they still fail to address the core representational challenge: handling entities with ambiguous references and resolving their mentions over time. Moreover, most common ground tracking research~\cite{ingress, williams, Madge2025ReferentialAA} has been conducted on the immediate conversational frame as compared to grounding over multiple events. \citet{Wu2024LongMemEvalBC} offers a dataset for multi-session dialogs but is devoid of complex relational references of situational dialogs.  


With the increasing use of LLMs in spoken dialog systems, their finite context window poses a constraint on maintaining long-term conversational memory. A common workaround is dialog summarization, where models generate concise summaries of past exchanges to be injected into future contexts, enabling a personalized experience~\cite{Ramprasad2024AnalyzingLB}. A more scalable solution involves Retrieval-Augmented Generation (RAG)~\cite{lewis-et-al}, which offloads the entire dialog history to external memory. During conversation, relevant excerpts are retrieved and fed into the LLM's context, a technique used to maintain long-term persona memory in conversational agents~\cite{shuster-etal-2021}. However, they have not been tested on their ability to resolve complex relational references. Consequently, an effective common ground representation must maintain information that is efficiently retrievable, dynamically updatable, and capable of tracking participant perspectives.

%% file: sections/metric.tex
\section{RQ1 : Benchmarking}

To measure grounding beyond superficial acknowledgment patterns, we turned to how humans leverage common ground. Inspired by \citet{kruijt-vossen-2022-role}, who showed that people frequently use relational references to access shared knowledge, we propose that a model's ability to resolve such references can serve as a robust metric for a model's grounding capabilities, one that can deal with dialogs where entities may not have a single, unambiguous referring expression.


To better understand the mechanisms of relational reference, we analyzed two dialog datasets previously used in grounding research. The Meetup dataset~\cite{meetup} places two participants on a 2D grid and tasks them with finding each other despite only seeing their own locations. Success in this setting depends on describing rooms, maintaining a record of shared knowledge, modeling the partner’s perspective, and clarifying ambiguities, all of which are central challenges for grounding. In contrast, the Spot the Difference~(STD)~\cite{spotthedifference} dataset involves two speakers, each given slightly different images, who must coordinate over an audio channel to identify discrepancies. While Meetup provides a scenario closer to real world situations for building common ground and referring back to it, we also used STD as it provides a different context to help test the robustness of our representation technique. Meetup corpus contains 5131 utterances, and Spot the difference contains 54 dialogs with 4934 utterances.

\input{tables/frequency}

We manually analyzed each utterance in the dialog datasets to identify references to previously mentioned entities (details in Appendix \ref{apx:analysis}). This analysis yielded a taxonomy of four primary categories: 1) \textbf{Temporal}, referring to an entity by its order of appearance (e.g., "\textit{the Thai restaurant that we went to after watching Spiderman}"); 2) \textbf{Spatial}, locating an entity relative to another (e.g., "\textit{the bottle on the table}"); 3) \textbf{Attributive}, describing an entity by its intrinsic properties (is-a: "\textit{yellow house}" or has-a: "\textit{sofa room}"); and 4) \textbf{Comparative}, using a relative property (e.g., "\textit{bigger chair}"). Furthermore, we observed implicit grounding mechanisms akin to the "indirect acknowledgments" identified by \citet{Mohapatra2024ConversationalGA}. We termed them \textbf{inferred grounding}, where participants ground information that was implied rather than explicitly stated. For instance, when a navigator is asked by their partner whether they see a "\textit{pink towel}" in their bathroom, that navigator also grounds the fact that their partner has visited a bathroom with a pink towel, later referring to it as "\textit{the bathroom with the pink towel that you had seen}". The frequencies found from our analysis are presented in Table~\ref{Table:freq}. Given the general difficulty of isolating specific phenomena within a limited number of spoken dialog corpora~\cite{sparsity, phenomena}, the identification of multiple occurrences in these two small datasets~(though limited in number) offers evidence for the use of `relational references' to refer back to established common ground. 

\vspace{-1mm}
\input{tables/llm_result}

\subsection{\indiref \ data creation}
\vspace{-1mm}
Our goal here is to extend beyond prior work to evaluate a model's ability to establish a persistent, useful and usable common ground. This capability becomes critical given the finite context windows of LLMs, which necessitate the efficient storage and retrieval of shared information. To this end, we introduce \indiref, a benchmark consisting of Question/Answer (Q/A) pairs where each question seeks information regarding an entity that is identified via a relational reference.

We focus on three primary reference types based on their conversational frequency in Table~\ref{Table:freq}: \textbf{Temporal}, \textbf{Spatial}, and \textbf{Attributive}. We also used the fourth category \textbf{inferred grounding}, to specifically probe the model's understanding of implicit grounding. As noted previously (Table~\ref{Table:freq}), the natural density of these references in the Meetup and STD datasets is low; therefore, we manually constructed 400 Q/A pairs (100 per category) to ensure sufficient coverage and unambiguous answers.

\textbf{Robustness:} \indiref \ is designed to be adversarial to superficial heuristics. By taking dialogs containing multiple entities of the same type (e.g., containing several `sofas'), we force the model to use the provided contextual reference (e.g., "\textit{color of the sofa in the \textbf{tv room}}" vs "\textit{color of the sofa in the \textbf{room in front of the bathroom}}") to disambiguate the target entity rather than relying on simple keyword matching. A correct answer thus serves as strong evidence of successful reference resolution, indicating that the model was able to ground the information properly in the first place. 

\textbf{Perspective}: Additionally, the benchmark evaluates perspective-taking through the use of deictics~\cite{diectic}. Questions are framed as the final utterance of a participant, requiring the model to respond as the interlocutor. This tests the model's ability to distinguish between ``self" and ``other" within the shared context. For instance, correctly answering "\textit{What was the color of the sofa in the second house that I visited?}" requires the model to recall the information shared by the questioner about their sequence of actions. However, this changes fundamentally if the question replaces `\textit{I}' with `\textit{you}', which would require the model to access shared information pertaining to its own character's experiences, thereby testing its ability to maintain distinct perspectives within the common ground. 


The multi-dimensional nature of \indiref \ provides a crucial metric to guide the development of models capable of creating more robust and usable common ground. For further details on benchmark construction, illustrative examples, and our human validation study that yielded an accuracy of 87.5\%, please refer to Appendix \ref{apx:test_example}. The benchmark can be found in \href{https://osf.io/92q4w/overview?view_only=88755344e00a401eb160b82227b1c099}{this link}.

%% file: tables/frequency.tex
\begin{table}
\centering
\resizebox{0.75\columnwidth}{!}{%
\begin{tabular}{|c|cc|}
\hline
\multirow{2}{*}{\begin{tabular}[c]{@{}c@{}}\textbf{\small{Phenomena}}\end{tabular}} & \multicolumn{2}{c|}{\textbf{\small{Dataset}}}                                           \\ \cline{2-3} 
                                                                                 & \multicolumn{1}{l|}{\small{Meetup}} & \multicolumn{1}{l|}{\small{Spot the difference}} \\ \hline
\small{Temporal Reference}                                                                         & \multicolumn{1}{c|}{\small{5}}      & \small{4}                                        \\ \hline
\small{Spatial Reference}                                                                          & \multicolumn{1}{c|}{\small{34}}     & \small{72}                                       \\ \hline
\small{Comparative Reference}                                                                      & \multicolumn{1}{c|}{\small{0}}      & \small{6}                                        \\ \hline
\small{Attributive Reference}                                                                       & \multicolumn{1}{c|}{\small{33}}     & \small{8}                                        \\ \hdashline
\small{Inferred grounding}                                                                         & \multicolumn{1}{c|}{\small{177}}    & \small{67}                                       \\ \hline
\end{tabular}
}
\caption{Frequency of different reference types}
\label{Table:freq}
\vspace{-5mm}
\end{table}

%% file: tables/llm_result.tex
\begin{table*}[htbp]
\centering
\resizebox{0.75\linewidth}{!}{%
\begin{tabular}{lcccccccc}
\toprule
\multirow{2}{*}{LLMs} & \multicolumn{4}{c}{Meetup} & \multicolumn{4}{c}{Spot the difference} \\ 
\cmidrule(lr){2-5} \cmidrule(lr){6-9}
                      & Temporal  & Spatial & Attributive & Inferred & Temporal & Spatial & Attributive & Inferred \\ \midrule
Gemma2 - 2B           &  0.20 / 0.18    &  0.18 / 0.16       &  0.24 / 0.26          &  0.26 / 0.16       &  0.18 / 0.12         & 0.18 / 0.10        &  0.20 / 0.12          &  0.16 / 0.10       \\ \midrule
Llama3.1 - 8B         &  0.38 / 0.32        &  0.46 / 0.38       &   0.46 / \textbf{0.44}         &  0.20 / 0.20       & 0.40 / \textbf{0.38}         &  \textbf{0.30} / \textbf{0.32}       &  \textbf{0.36} / \textbf{0.36}          &  0.26 / 0.16   \\ \midrule
Gemma2 - 9B           &  0.48 / 0.38         &   0.50 / 0.46     &  0.38 / 0.40          &  0.26 / 0.18       & 0.34 / 0.20        &  0.24 / 0.16       & 0.28 / 0.16           & 0.20 / 0.10        \\ \midrule
Gemma2 - 27B          &  \textbf{0.50} / \textbf{0.44}         &  \textbf{0.58} / \textbf{0.56}       &  \textbf{0.48} / \textbf{0.44}          &   0.28 / 0.26      &  0.40 / 0.30        &  0.28 / 0.20       &   0.32 / 0.26         &   0.26 / 0.12      \\ \midrule
Qwen QwQ - 32B        &   0.38 / 0.32        &   0.52 / 0.38      &  0.44 / 0.40          &   \textbf{0.40} / \textbf{0.40}      & \textbf{0.42} / 0.34         & 0.26 / 0.22        &  0.26 / 0.20          &  \textbf{0.28} / \textbf{0.20}      \\ \bottomrule
\end{tabular}
}
\caption{ FEM~($\uparrow$) / LLM-as-Judge~($\uparrow$) accuracy for baselines where the entire dialog history is used as the context to answer the question. Best results in \textbf{bold}.}
\label{Table:llm_result}
\vspace{-3mm}
\end{table*}

%% file: sections/baseline.tex
\input{tables/rep_baseline}

\section{RQ2 : Representation}
\subsection{Baseline - Full Dialog Context}
\vspace{-1mm}

To evaluate common ground representation capabilities, we adopt a framework inspired by \cite{survey_memory} utilizing three operators: Writer ($W$), Reader ($R$), and Generator ($G$). As a baseline, we assess the LLM's internal ability to ground context by providing the full dialog history ($h_t$) (containing the speaker information, utterances, and the time stamps) and tasking them with generating a response ($u_{t+1}$) to a question in the preceding utterance. We test a range of open-weight instruction-tuned models: Gemma2 (2B, 9B, 27B)\cite{gemma2}, Llama 3.1 8B\cite{llama3}, and reasoning-focused Qwen-QWQ-32B~\cite{qwenqwq}. In this setup, $W$ appends each new utterance to the history, $R$ provides the entire history to the model, and $G$ produces the subsequent utterance.

\vspace{-4mm}
\begin{small}
    \begin{equation*}
        W(h_{t-1}, u_{t-1}) = h_t; \; R(u_{t})=h_{t}; \; G(h_t, u_{t}) = u_{t+1}
    \end{equation*}
\end{small}

\vspace{-3mm}
\subsection{Resource Constrained Setting}


As dialogs extend over time, retaining the complete interaction history becomes prohibitive in terms of context window, computation and inference latency.
Although the \indiref \ dataset fits within modern LLM context windows, we artificially restrict context access to simulate the `longer context' scenarios required for persistent agents.
Despite lacking extreme temporal length, the underlying corpora of \indiref, particularly the Meetup dataset, offer a critical testing ground because of their fragmented context i.e., speakers discuss and recall information exchanged across different contexts(rooms in this case) as the conversation progresses. Under our restricted context settings, this dynamic effectively simulates the episodic transitions of long-term interactions, forcing the model to resolve references that lie outside the immediate context window.

We compare three commonly used common ground representation methods under these constraints.

\textbf{Summarization:} A method shown by \citet{padmakumar-etal-2023-investigating} to be an effective representation of open-dialog context for transformer-based models. Its utility is further underscored by its adoption in modern LLM chatbots for personalization~\cite{Zhang2024PersonalizationOL} by summarizing each session separately. In this scheme, $W$ creates a summary ($s_t$) of the whole dialog, which is stored, retrieved by $R$, and used by $G$ to form the next utterance.

\vspace{-3mm}
\begin{small}
    \begin{equation*}
        W(s_{t-1}, u_{t-1}) = s_t; \; R(u_{t})=s_t; \; G(s_t, u_{t}) = u_{t+1}
    \end{equation*}
\end{small}

\vspace{-1mm}
\textbf{Utterance Chunking:} Our first method to explicitly group information to enhance retrieval. Here, $W$ segments the dialog into overlapping chunks of utterances ($c_i$). For our experiments, we define a chunk as seven utterances with a stride of three to ensure informational continuity. $R$ then retrieves the top three chunks most relevant to the user's utterance by matching its embedding against the embeddings of each chunk. To isolate the performance of this structural approach, our Reader only uses the embeddings of the concatenated utterances within a chunk, and does not use additional embeddings of chunk summaries as done by~\cite{Wu2024LongMemEvalBC}. To better understand how embeddings affect the retrieval of important context, we tested two types of embedders: a sparse model (BM25~\cite{bm25}) and a dense model (NV-Embed-V2~\cite{lee2024nv}). 

\vspace{-2mm}
\begin{minipage}{0.7\columnwidth}
\begin{small}
\begin{flalign*}
& W(h_{t-1},u_t) = [c_1,\dots,c_i] &\\
& R(u_{t+1}) = \operatorname{top}_k\!\bigl(\mathrm{sim}(e(u_{t+1}),[e(c_1),\dots,e(c_i)])\bigr) &\\
& G(u_t,[c_1,\dots,c_k]) = u_{t+1} &
\end{flalign*}
\end{small}
\end{minipage}

\vspace{2mm}
\textbf{Ontology:} We use an \textbf{agentic} approach which investigates the use of an \textbf{ontology} ($O$) to create a structured common ground. The Writer ($W$) processes dialog chunks to extract key entities, assigning each a unique id and mapping its properties, inter-entity relations and source speaker as a proxy for estimating each speaker's belief. This rich structure enables complex relational reasoning (e.g., if \textit{`place A has object b'} and \textit{`b is kept on top of c'}, then \textit{`A also has c'}). To preserve temporal context, $W$ also maintains a chronological log of events, which includes the involved entity IDs and speakers, serving as a primary index for querying. Given this highly structured knowledge, $R$ uses a multi-step query process to retrieve and process information before sending it to $G$. It first decomposes the user's request into a series of sub-queries~(q), each prefixed with a command: $RAG[n]$, $Process$, or $Final$. A $RAG[n]$ query retrieves the top-n relevant contexts~($c$) from the extracted common ground. Here n is also generated by the LLM. A $Process$ query then performs operations on this retrieved data, such as filtering or selecting specific items. The final retrieved context and the $Final$ query are passed to $G$ to synthesize the final answer. All operations are implemented using langchain~\cite{langchain} for the vectorstore and NV-Embed-V2 for embedding generation.

\vspace{-4mm}
\begin{small}
    \begin{align*}
        W(O_{t-1}, u_{t-1}) &= O_t\\
        R(u_{t}) &= [q_1....q_m] \\
        R(q_i, O_t, c_{i-1}) &= c_i \\
        G(q_m, c_{m-1}) &= u_{t+1}
    \end{align*}
\end{small}

%% file: tables/rep_baseline.tex
\begin{table*}[htbp]
\centering
\resizebox{0.75\linewidth}{!}{%
\begin{tabular}{lcccccccc}
\toprule
\multirow{2}{*}{LLMs} & \multicolumn{4}{c}{Meetup} & \multicolumn{4}{c}{Spot the difference} \\ 
\cmidrule(lr){2-5} \cmidrule(lr){6-9}
                      & Temporal  & Spatial & Attributive & Inferred & Temporal & Spatial & Attributive & Inferred \\ \midrule
Full Dialog [from Table~\ref{Table:llm_result}]      & 0.38 / 0.32        &  \textbf{0.46} / \textbf{0.38}       &   \textbf{0.46} / \textbf{0.44}         &  0.20 / 0.20       & \textbf{0.40} / \textbf{0.38}         &  \textbf{0.30} / \textbf{0.32}       &  \textbf{0.36} / \textbf{0.36}          &  \textbf{0.26} / \textbf{0.16}   \\ \midrule
Summarisation           &  0.32 / 0.28      & 0.34 / 0.26        &  0.30 / 0.25          &  \textbf{0.28} / 0.18       &  0.22 / 0.16        &  0.24 / 0.16       &   0.34 / 0.26         &   0.20 / 0.12      \\ \midrule
Chunking (NV-Embbed-V2)         &  0.24 / 0.20         &   0.08 / 0.06      &  0.16 / 0.08          &   0.22 / 0.24      &   0.02 / 0.02       &  0.04 / 0       &    0.06 / 0.02        &   0.12 / 0.08      \\ \midrule
Chunking (BM25)          &  0.26 / 0.24         &   0.20 / 0.16      &    0.20 / 0.18        &  0.24 / \textbf{0.26}       &   0.04 / 0       &  0 / 0       &   0.10 / 0.08         &   0.12 / 0.10      \\ \midrule
Agentic Ontology-based       &  \textbf{0.40} / \textbf{0.36}         &  0.38 / 0.34       &    0.38 / 0.30        &   0.24 / 0.22      &  0.20 / 0.14        &  0.12 / 0.14       &  0.20 / 0.24          &  0.18 / 0.14      \\ \bottomrule
\end{tabular}
}
\caption{FEM~($\uparrow$) / LLM-as-Judge~($\uparrow$) accuracy for various representations using Llama 3.1 8B. Best in \textbf{bold}.}
\label{Table:rep_baseline_result}
\vspace{-4mm}
\end{table*}

%% file: sections/results.tex
\subsection{Evaluation}

To comprehensively evaluate performance, we employ a Fuzzy Exact Matching~(FEM) metric. For each question, an answer is considered correct if it contains a predefined keyword or phrase. Recognizing that keyword presence alone may not guarantee accuracy, we also incorporate an LLM-as-a-judge approach~\cite{judge}. A Qwen2.5-32B-Instruct~\cite{qwen2.5} model, provided with the question and both the correct and generated answers, determines correctness based on semantic equivalence. Crucially, dialog context was intentionally excluded from the LLM's input to ensure its judgment focuses solely on the answers' semantic similarities. Since the relational references point to only one correct referent, the overall performance for both fuzzy matching and the LLM-as-a-judge approach was quantified using accuracy. To validate the reliability of our automated judge, we conducted a human evaluation on a subset of answer pairs. We observed a 90\% agreement rate, demonstrating a strong alignment between human and LLM judgments (details in Appendix \ref{apx:irr_llm}).

Table~\ref{Table:llm_result} presents the performance of various LLMs when provided with the full dialog context. For Meetup, results indicate that model performance improves with increasing size, with Gemma2-2B consistently exhibiting the lowest accuracies across all categories. The relative performance of Fuzzy Matching and LLM-as-a-judge remains consistent across models. Slight variations occur when Fuzzy Matching scores differ due to longer answers with multiple entities or alternate representations (e.g., digits vs. words). 
Surprisingly, Gemma2-27B emerges as the top performer across all evaluated models, outperforming even Qwen-QWQ which is a slightly larger model specifically trained for enhanced reasoning. Further analysis of Qwen-QWQ's responses revealed difficulties in logical reasoning and frequent hallucinations(Appendix; Figure~\ref{fig:Qwen_wrong}), potentially attributable to a scarcity of spoken dialogs in its training data. Despite this, Qwen-QWQ excelled in inferred grounding questions, a category where other models performed poorly, underscoring the critical role of reasoning for grounding implicit information. It also had better fuzzy exact match scores due to its verbose nature. However, in STD, Gemma models underperformed compared to Llama. We hypothesis that the longer dialog context in STD as compared to Meetup helped Llama which uses RoPE over Gemma which uses sliding window attention. Furthermore, the models demonstrated significant limitations in differentiating between speaker perspectives (e.g., `your' vs `mine') and in isolating information conveyed by individual speakers, often resulting in the erroneous merging of conversational content. It is critical to note that \textbf{none of the baselines achieved an accuracy exceeding 50\% across all the categories}.

\input{tables/training_full_dialog}

\input{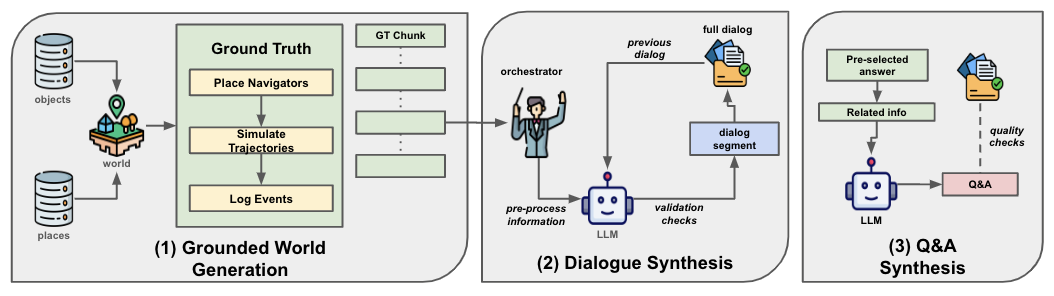}

Table~\ref{Table:rep_baseline_result} details the performance of various resource-constrained representation techniques. Due to space constraints, we primarily report results for the Llama3.1-8B model; results for the Gemma2 2B and 9B models are provided in Appendix~\ref{apx:test_result}. The general trend across all models remained consistent. We did not test larger models due to resource limitations. All the representation techniques yielded poorer performance compared to the baseline, a likely consequence of information loss during the extraction of relevant information. We noticed that sparse embedding (BM25) demonstrated a slight advantage over dense embedding, which may be attributed to its enhanced capability in named-entity retrieval. The agentic ontology-based system outperformed other resource-constrained representation techniques. While it performed relatively well on the Meetup dataset, it struggled with the STD dataset, often merging the representations of images of both participants due to their similarity. However, its better performance suggests that a multi-step retrieval process, particularly one that explicitly extracts speaker-specific information regarding entities and events, helps the LLM understand the context better. \href{https://github.com/biswesh456/Representation-and-storage-of-grounded-information}{Code link for evaluation}.

%% file: tables/training_full_dialog.tex
\begin{table*}[htbp]
\centering
\resizebox{0.8\linewidth}{!}{%
\begin{tabular}{lcccccccc}
\toprule
\multirow{2}{*}{LLMs} & \multicolumn{4}{c}{Meetup} & \multicolumn{4}{c}{Spot the difference} \\ 
\cmidrule(lr){2-5} \cmidrule(lr){6-9}
                      & Temporal  & Spatial & Attributive & Inferred & Temporal & Spatial & Attributive & Inferred \\ \midrule
Original Setting [Table~\ref{Table:llm_result}]          &  0.38 / 0.32        &  0.46 / 0.38       &   0.46 / 0.44         &  0.20 / 0.20       & 0.40 / 0.38         &  0.30 / 0.32       &  0.36 / 0.36          &  0.26 / 0.16  \\ \hdashline
In-Context Learning         &  \textbf{0.60} / \textbf{0.56}         & 0.58 / \textbf{0.54}        &   \textbf{0.62} / 0.58         &   0.42 / 0.34      &  0.48 / 0.52        &  \textbf{0.48} / 0.40       &   0.52 / 0.54         &  0.36 / \textbf{0.44}       \\ \midrule
GRPO Trained       &  0.58 / 0.52        &  \textbf{0.66} / \textbf{0.54}       & \textbf{0.62} / \textbf{0.60}          & \textbf{0.46} / \textbf{0.42}       &   \textbf{0.52} / \textbf{0.54}       &  \textbf{0.48} / \textbf{0.42}      &   \textbf{0.56} / \textbf{0.58}        &  \textbf{0.38} / 0.42      \\ \bottomrule
\end{tabular}
}
\caption{FEM~($\uparrow$) / LLM-as-Judge~($\uparrow$) accuracy on entire dialog history for enhanced Llama 3.1-8B. Best in \textbf{bold}.}
\label{Table:full_dialog_result}
\vspace{-3mm}
\end{table*}

%% file: figures/data_gen.tex
\begin{figure*}[ht]
  \centering
  \includegraphics[width=0.75\linewidth, height=3.22cm]{
  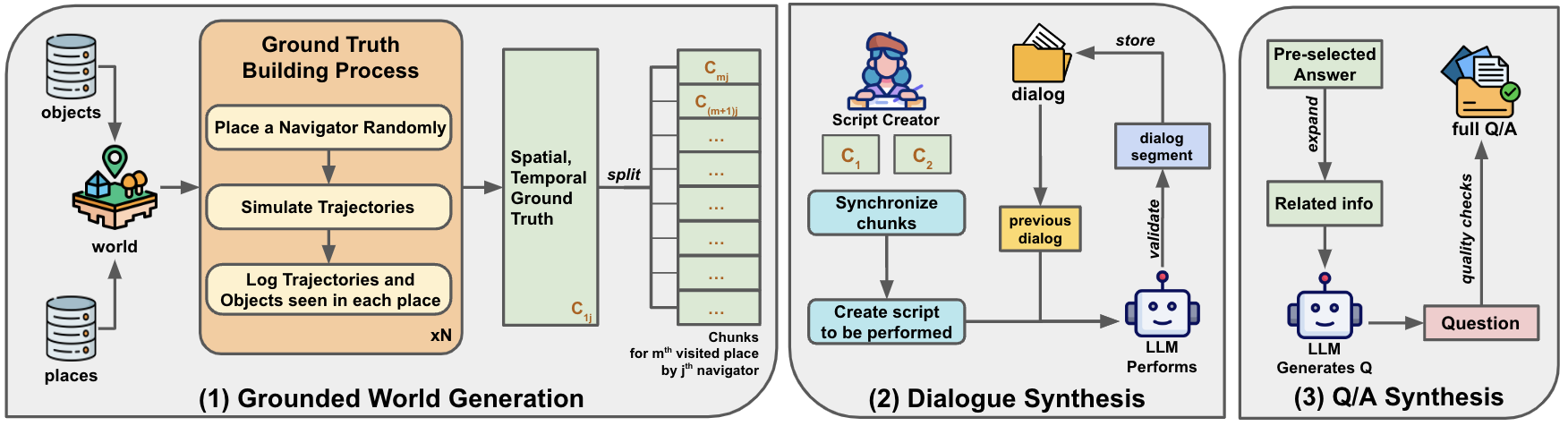}
\caption{Pipeline for the synthetic data generation involving three main phases: (1) Grounded World Generation, creates the factual basis of the scenario; (2) Dialog Synthesis, where LLM generates conversation constrained by the script; and (3) Q/A Synthesis, which extracts and formats question-answer pairs for training.}
\label{fig:data_gen}
\vspace{-5mm}
\end{figure*}

%% file: sections/improve.tex
\section{RQ3 : Improving Performance}

Error Analysis of the thinking tokens from baseline models indicated that they primarily struggled to link references to entities requiring multi-hop reasoning. Despite general advancements in reasoning capabilities, our findings showed that LLMs continue to face significant difficulties within situated dialog settings. Further, a significant challenge lies in their understanding of overlapping information from multiple speakers, preventing the construction of a reliable grounding representation. Thus, our initial research focused on enhancing the performance of baseline models leveraging the entire dialog history. We implemented In-Context Learning~\cite{incontext} by augmenting the input prompts with examples for each reference type to instruct the LLMs on how to accurately ground the context within the dialogs. The prompts~(see Appendix~\ref{apx:prompts}) were further enriched with explicit guidance on decoupling the information contributed by each speaker, unless a common entity was being discussed. As shown in Table~\ref{Table:full_dialog_result}, this resulted in a significant improvement in LLM performance across all categories in both datasets. These findings indicate that enhanced prompt design can substantially improve the models' comprehension of spoken conversations, a critical prerequisite for effective grounding.

To enable the model to develop inherent grounding capabilities applicable to diverse settings, we used reinforcement learning to improve the model's multi-hop reasoning capability. We used Group Relative Policy Optimization~(GRPO)~\cite{grpo} to train the model to identify the important context by providing positive rewards for each accurate response. A primary challenge was the absence of suitable training data for situational spoken dialogs. Thus, we generated synthetic data to resolve data scarcity. This approach not only resolved data scarcity but also facilitated arbitrary scaling, fine-grained control over domains, and improved coverage of edge cases.
\input{tables/ontology}
\vspace{-2mm}
\subsection{Synthetic Data}
Given that off-the-shelf LLMs lack proficiency in spoken dialog, we designed a systematic, multi-phase synthetic data generation framework with an "environment-first, dialog-later" approach. Inspired by procedural content generation~\cite{smith2015analog} and recent advances in synthetic dialog generation~\cite{simulated-chats, suresh2024diasynth}, our framework is engineered to produce a grounded situated dialog corpus. The process is structured into three phases as shown in Figure~\ref{fig:data_gen}:

\textbf{Grounded World Generation}: First, a rich and verifiable simulated ‘world’ is programmatically constructed, starting from a list of places and related objects previously generated (see Appendix \ref{apx:pipeline}). Two algorithmic navigators explore this world, and their experiences are logged to establish the spatio-temporal ground truth (i.e., what they saw, where they went, etc.) that we can then split into smaller chunks, where $\mathcal{C}_{m,j}$ is the chunk for the $m^{th}$ visited place by the $j^{th}$ navigator.

\textbf{Dialog Synthesis}: We then synthesize dialogs where the two navigators describe their locations to find each other. Dialogs are constrained by the rules and facts of the created world. A programmatic Script Creator synchronizes the chunks of the two Navigators and generates the script for the LLM to follow, thus generating dialog segments based on the event logs. In this way, we rely on the Script Creator for the ‘reasoning’ part of the task (i.e., noticing similarities, addressing ambiguities, resolving references, etc), while the LLM focuses on constrained Utterance Generation. This generation process is iterative and considers the previously generated utterances for cohesiveness.

\textbf{Q/A Synthesis:} The final stage automatically identifies scenarios to generate accurate Question-Answer (Q/A) pairs requiring complex reference resolution. The process is deterministic: a fact is preselected from the Ground Truth as the ‘expected answer’, and the LLM is prompted to generate a question, constrained to the context, that elicits this specific answer. These are designed to probe various aspects of common ground, including spatial, temporal, and relational facts, and undergo a strict validation process. We generated $\sim600$ Q/A pairs from various synthetic dialog scenarios.

Table~\ref{Table:full_dialog_result} demonstrates that reinforcement learning on the synthetic data improved model accuracy by 15-20\% not only on the Meetup dataset (which closely resembles the training data), but also on the Spot-the-Difference dataset. This indicates that the models successfully acquired the underlying concepts needed to resolve such relational references to the common ground. We note that for the incorrect responses, the models would often reason in the right direction but then hallucinate while providing the final answer. We expect this to improve with larger models but, due to lack of computational resources, we did not train on them.

To assess whether the enhanced contextual understanding of GRPO-trained models improves their representational capabilities, we applied the agentic ontology-based representation using the same GRPO-trained model trained on synthetic Q/A pairs, given its promising results in earlier experiments. As shown in Table~\ref{Table:ontology}, the GRPO-trained model outperformed its non-GRPO counterpart on the Meetup dataset but continued to struggle with the Spot-the-Difference dataset. The main issue concerned representing the two images as distinct within the common ground, as the models frequently merged them due to their high degree of similarity. While more carefully designed task-specific prompts could improve performance, these would limit generalizability. These findings therefore suggest that better multi-step reasoning alone does not guarantee a robust, reusable representation of persistent common ground. Here is the \href{https://github.com/r3lativo/synth4indiref}{link} for the synthetic data generation code.


\input{figures/agentic_rl}
\input{tables/rl_agent}

\subsection{RL for optimizing the externalized common ground representation}

We further investigated whether RL could directly optimize the \emph{externalized} common ground representation produced by our agentic framework. In this setting, the common ground generator~(Writer) serves as the policy: given a dialog, it produces a common ground representation that is then consumed by the Reader and Generator modules to answer an IndiRef question. The correctness of the final answer, as determined by our evaluation pipeline, was used as the reward signal for the representation step (see Figure~\ref{fig:agentic_rl}). Training was conducted on the synthetic dialogs introduced above.

We evaluated two variants. The \textbf{template constrained} setting required the model to populate our ontology-style schema. This meant that the overall structure of the representation was fixed in advance, and reinforcement learning could only optimize \emph{what} information to include within that structure, such as which entities, attributes, relations, and temporal events were most useful for later retrieval. In effect, this setup tests whether RL can make a predefined representation more functionally informative. The \textbf{open-ended} setting removed the schema entirely and asked the model to generate any representation that would allow another model to answer future questions from the shared context. Here, the model was not given an explicit ontology or field structure, and therefore had to decide both \emph{what} information to preserve and \emph{how} to organize it. This setup tests whether the model can discover, through reward alone, a representation strategy better suited for downstream retrieval than a hand-designed template.

Contrary to our expectations, both RL settings underperformed the off-the-shelf ontology pipeline (Table~\ref{Table:rl_agent}). The template-constrained variant did not improve over the original system, and the open-ended variant collapsed to near-zero accuracy. Our error analysis suggests that the main bottleneck is the \textbf{credit assignment problem} in multi-step agentic pipelines. Since reward is assigned only from the correctness of the final answer, it does not cleanly reflect the quality of the generated common ground. As a result, training was corrupted by \textbf{Spurious Rewards}, in which poor representations were reinforced because downstream modules still produced the correct answer, and \textbf{Unjustified Penalties}, in which high-quality representations were discouraged because later stages failed to use them correctly. In the open-ended setting, the absence of structural constraints led the model to produce generic summaries rather than the fine-grained relational content needed for IndiRef.

These findings indicate that optimizing externalized common ground via end-to-end RL is substantially harder than improving direct answer generation. They also suggest that for grounding representation, a structured template provides an important inductive bias that cannot be reliably replaced by open-ended generation.



%% file: tables/ontology.tex
\begin{table*}[htbp]
\centering
\resizebox{0.8\linewidth}{!}{%
\begin{tabular}{lcccccccc}
\toprule
\multirow{2}{*}{LLMs} & \multicolumn{4}{c}{Meetup} & \multicolumn{4}{c}{Spot the difference} \\ 
\cmidrule(lr){2-5} \cmidrule(lr){6-9}
                      & Temporal  & Spatial & Attributive & Inferred & Temporal & Spatial & Attributive & Inferred \\ \midrule
Without GRPO training [Table~\ref{Table:rep_baseline_result}]          &  0.40 / 0.36         &  0.38 / 0.34       &    0.38 / 0.30        &   0.24 / 0.22      &  0.20 / 0.14        &  0.12 / 0.14       &  0.20 / 0.24          &  0.18 / 0.14   \\ \hdashline
With GRPO training       &  0.48 / 0.46         & 0.44 / 0.42       & 0.52 / 0.44          & 0.36 / 0.38       &   0.20 / 0.24      &  0.22 / 0.28      &  0.28 / 0.32         &     0.16 / 0.20  \\ \bottomrule
\end{tabular}
}
\caption{FEM~($\uparrow$) / LLM-as-Judge~($\uparrow$) accuracy for agentic ontolgy-based representation after GRPO training.}
\label{Table:ontology}
\vspace{-4mm}
\end{table*}

%% file: figures/agentic_rl.tex
\begin{figure*}[h]
  \centering
  \includegraphics[width=0.65\linewidth, height=40mm]{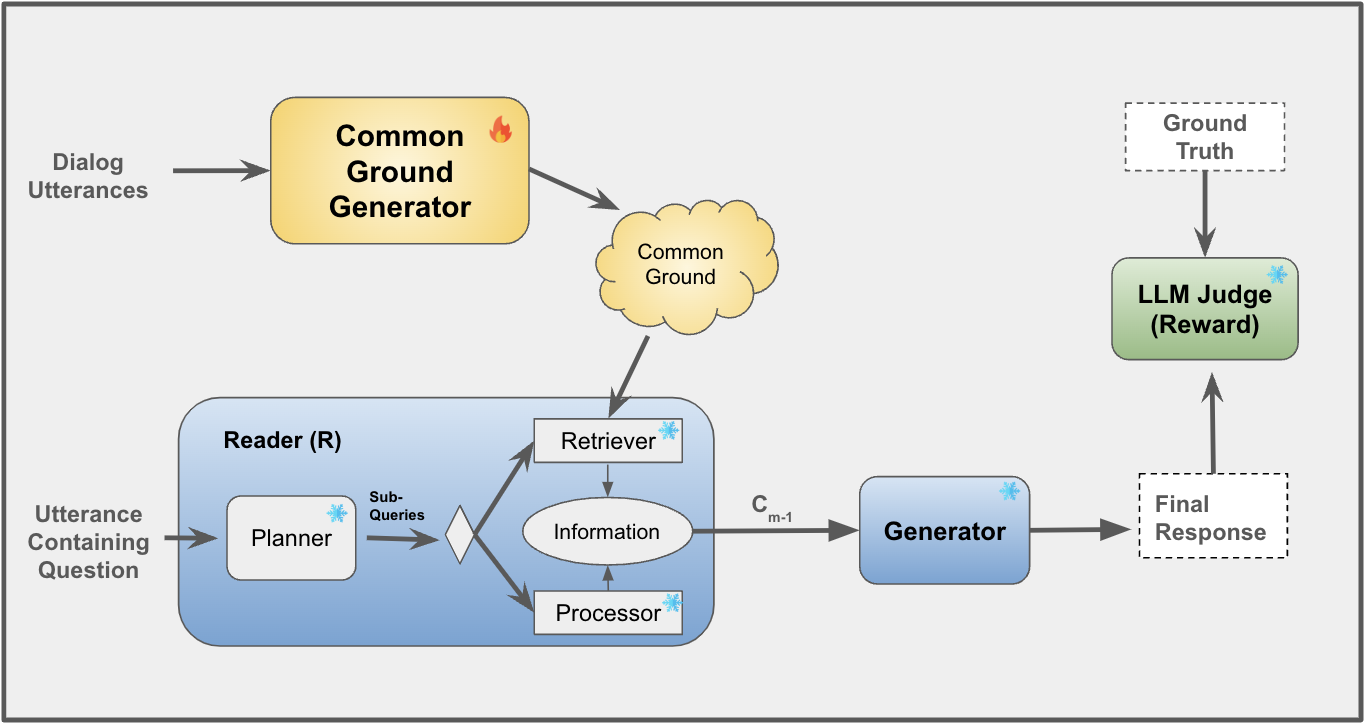}
\caption{Visual representation of RL for optimizing the externalized common ground representation.}
  \label{fig:agentic_rl}
\end{figure*}

%% file: tables/rl_agent.tex

\begin{table*}[htbp]
\centering
\resizebox{0.8\linewidth}{!}{%
\begin{tabular}{lcccccccc}
\toprule
\multirow{2}{*}{Setting} & \multicolumn{4}{c}{Meetup} & \multicolumn{4}{c}{Spot the Difference} \\
\cmidrule(lr){2-5} \cmidrule(lr){6-9}
 & Temporal & Spatial & Attributive & Inferred & Temporal & Spatial & Attributive & Inferred \\
\midrule
Off-the-shelf Llama [Table~\ref{Table:rep_baseline_result}]
& 0.40 / 0.36 & 0.38 / 0.34 & 0.38 / 0.30 & 0.24 / 0.22
& 0.20 / 0.14 & 0.12 / 0.14 & 0.20 / 0.24 & 0.18 / 0.14 \\
\hdashline
Template Constrained
& 0.30 / 0.22 & 0.28 / 0.26 & 0.18 / 0.20 & 0.20 / 0.16
& 0.12 / 0.10 & 0.10 / 0.10 & 0.18 / 0.16 & 0.14 / 0.12 \\
Open Ended
& 0.04 / 0.04 & 0.02 / 0.04 & 0.06 / 0.06 & 0.02 / 0.02
& 0.02 / 0.04 & 0.04 / 0.04 & 0.08 / 0.06 & 0.00 / 0.00 \\
\bottomrule
\end{tabular}
}
\caption{FEM~($\uparrow$) / LLM-as-Judge~($\uparrow$) accuracy for agentic ontology-based representation where the common ground generator is directly trained using RL.}
\label{Table:rl_agent}
\vspace{-4mm}
\end{table*}

%% file: sections/conclusion.tex
\section{Conclusion and Future Work}

In this work, we used relational references to measure the capability of a dialog system to establish a persistent and useful common ground by developing the \indiref \ benchmark. While our baselines with full dialog context struggled to consistently achieve an accuracy of 50\%, models mimicking resource-constrained settings performed even worse. Our experiments showed that both in-context learning and GRPO training with synthetic data improved an LLM's ability to resolve relational references from the full dialog history.  The same GRPO-trained models moderately also enhanced common ground extraction for long-term use in agentic setups due to better multi-step reasoning. However, inconsistent performance highlight the need for more robust representation methods. For future work, while our study focused on interpretable textual representations of the common ground, exploring more efficient parametric methods is a promising direction. We believe that our benchmark and proposed methods provide a solid foundation for advancing research into more robust common ground representation in dialog agents.

%% file: sections/appdx.tex
\section{Conversational Data Analysis}
\label{apx:analysis}
The reference analysis in our conversational data was conducted through an iterative annotation process involving three independent raters with a background in Natural Language Processing. As we wanted to inductively discover the reference typology, we did not start with a pre-defined schema. Hence, the methodology was executed in multiple phases: We first took a few dialogs and each person marked the phrases of 'relational' references independently where we used the term 'relational reference' to group a reference to an entity or an event using associated contextual information, such as time, space, interlocutor identity etc. After this process, we did an Inter-rater reliability on the reference phrases discovered by the three annotators. On achieving a Krippendorff’s Alpha of 0.84, we annotated such reference phrases for the rest of the corpora. Once completed, we grouped the reference phrases manually and named each of the phrases. This led to the specific relational reference groups discussed in the paper i.e. Spatial, Temporal, Comparative, Attributive and Indirect.

\section{Benchmark Creation}
\label{apx:test_example}

The creation of our test cases involved an initial selection of 50 dialogs per category from both datasets. The authors of this paper then undertook the task of creating question/answer pairs that contained relational references to entities within the established common ground of each dialog. A key methodological consideration was the exclusive use of conversational data to formulate these questions. This approach ensures that the common ground is derived solely from the information discussed, rather than any external knowledge individual participants may have from the images they viewed. For validation, the benchmark was evaluated by 10 volunteers with expertise in Natural Language Processing. These volunteers were provided with the dialogs and asked to answer the corresponding questions. We randomly selected 10\% of test cases from each category for this validation process and each test case was provided to two volunteers. We then checked the accuracy of the answers from the human evaluators using the gold answers of the benchmark. The accuracy rate of 87.5\% affirmed the correctness of our question-answer pairs in the benchmark. Since no human evaluator pair received more than one question, we didn't calculate the inter-rater reliability between them. However, we would like to mention that we didn't find a single case where both the annotators got the answer wrong(i.e. different from gold answer). Thus the 87.5\% accuracy refers to the fact that the pairs agreed with each other on 87.5\% of the instances.
A subsequent error analysis indicated that the mistakes were a result of human error where they missed the mention of certain information.

Illustrative examples for each test case category are provided in Figures \ref{fig:test_ex_4}, \ref{fig:test_ex_2}, \ref{fig:test_ex_1}, and \ref{fig:test_ex_3}. Specifically, we present two examples (Spatial and Temporal) from the Spot the Difference dialogs and two examples (Attributive and Inferred) from the Meetup dialogs. In these figures, we have highlighted the essential segment of the dialog containing the answer (in \textcolor{purple}{purple}) to the question (in \textcolor{violet}{violet}). We also provide the keywords needed for the exact match analysis.

As can be seen in the examples, our questions are always regarding an object that has multiple instances with different properties. For example, Figure~\ref{fig:test_ex_4} provides an example of inferred grounding, where, while A asks B if his child's room has a teddy bear, B says no. This should ground two facts - A's room has a teddy bear while B's room doesn't.  However, in the question, A asked about his child's room. This should have made B answer about A's room but since in the original dialog, the answer to the question is `nope', our models generally provide a negative answer.
`...' in our figures indicate a time gap between the chunk of utterances as can be seen in Figure~\ref{fig:test_ex_3} where we have a gap from the time [02:21] till [10:45] and again till [12:04].
Figure~\ref{fig:test_ex_3} references a white cloud mentioned in preceding images, yet the question is specifically directed at the final image.

\begin{figure}[!htbp]
    \centering
    \begin{dialogueboxB}{Inferred} 
        \begin{dialoguelist}
            \item \texttt{[00:15]} \textbf{B}: hello
            \item \texttt{[00:29]} \textbf{B}: I've found a child's room.
            \item \texttt{[00:31]} \textbf{A}: i am in a child
            \item \texttt{[00:34]} \textbf{A}: 's room too
            \item \texttt{[00:42]} \textbf{A}: blue comforter and teddy bear?
            \item \texttt{[00:43]} \textbf{B}: nope.
            \item \texttt{[00:48]} \textbf{B}: There are two beds. with a purplish mattress, and a biege comforter across each
            \item \texttt{[00:57]} \textbf{A}: oh i can look for that one
            \item \texttt{[01:03]} \textbf{B}: okay.
            \item \texttt{[01:14]} \textbf{B}: there's also a set of pink and white towels hanging by the foot of the bed
            \item \texttt{[02:01]} \textbf{B}: there's a small white nightstand between them, with a little lamp on it.
            \item \texttt{[02:01]} \textbf{B}: I'm heading to a new location.
            \item \texttt{[02:01]} \textbf{A}: i am in the bathroom
            \item \texttt{[02:12]} \textbf{A}: where are you at?
            \item \texttt{[02:16]} \textbf{B}: I'm outside. Let's move around a bit.
            \item \texttt{[02:21]} \textbf{A}: k
            \item ...
            \item \textcolor{violet}{\texttt{[08:23]} \textbf{A}: Was there a teddy bear in that child's room that I visited?}
            \item \textcolor{purple}{\textbf{B}: Yes there was}
            \item \textcolor{purple}{\textbf{Exact Match}: Yes}
        \end{dialoguelist}
    \end{dialogueboxB}
    \caption{Test example for inferred grounding from Meetup dataset}
    \label{fig:test_ex_4}
\end{figure}

\begin{figure}[htbp]
    \centering
    \begin{dialogueboxB}{Attributive} 
        \begin{dialoguelist}
            \item ...
            \item \texttt{[02:37]} \textbf{A}: I am in the basement
            \item \texttt{[02:37]} \textbf{B}: I'm in a basement.
            \item \texttt{[02:49]} \textbf{B}: Mine has a white staircase
            \item \texttt{[02:54]} \textbf{A}: No
            \item \texttt{[03:03]} \textbf{A}: mine has wooden stair case
            \item \texttt{[03:21]} \textbf{B}: Mine has a leather chair. Should I try to move towards you?
            \item \texttt{[03:35]} \textbf{A}: Sure
            \item \texttt{[03:37]} \textbf{B}: Wooden? What else?
            \item \texttt{[03:53]} \textbf{A}: water heater and washer and dryer
            \item \texttt{[04:08]} \textbf{A}: a plastic chair and a screen door
            \item \texttt{[04:08]} \textbf{A}: I'm heading to a new location.
            \item \texttt{[04:08]} \textbf{A}: I have found a staircase
            \item \texttt{[04:34]} \textbf{A}: The staircase is on my right and there is a hallway in front of me
            \item \texttt{[04:58]} \textbf{A}: Find it?
            \item \texttt{[05:01]} \textbf{B}: found the staircase
            \item \texttt{[05:15]} \textbf{A}: is it the one with the hallway in front of you?
            \item \texttt{[05:27]} \textbf{B}: it is to my east
            \item \texttt{[05:30]} \textbf{A}: and the staircase on the right?
            \item \texttt{[05:38]} \textbf{B}: its on my left
            \item \texttt{[05:53]} \textbf{A}: We're not at the same staircase
            \item ...
            \item \textcolor{violet}{\texttt{[07:12]} \textbf{B}: What was the material of the chair present in the \textbf{white staircase basement}?}
            \item \textcolor{purple}{\textbf{A}: It was a leather chair.}
            \item \textcolor{purple}{\textbf{Exact Match}: leather}
        \end{dialoguelist}
    \end{dialogueboxB}
    \caption{Test example for attributive grounding from Meetup dataset.}
    \label{fig:test_ex_2}
\end{figure}

\begin{figure}[htbp]
    \centering
    \begin{dialogueboxB}{Spatial} 
        \begin{dialoguelist}
            \item ...
            \item \texttt{[09:45]} \textbf{A}: Okay I see.
            \item \texttt{[09:46]} \textbf{A}: Two sharks on the right top
            \item \texttt{[09:49]} \textbf{A}: of the image.
            \item \texttt{[09:50]} \textbf{A}: They are blue
            \item \texttt{[09:51]} \textbf{A}: and gray.
            \item \texttt{[09:52]} \textbf{A}: Or
            \item \texttt{[09:52]} \textbf{A}: another kind of blue.
            \item \texttt{[09:53]} \textbf{B}: The one below.
            \item \texttt{[09:55]} \textbf{A}: Yes.
            \item \texttt{[09:56]} \textbf{B}: In my
            \item \texttt{[09:57]} \textbf{B}: picture it's green, the one on the top is blue. Which color did you say it was yours?
            \item \texttt{[10:01]} \textbf{A}: They are both blue and
            \item \texttt{[10:03]} \textbf{A}: gray.
            \item \texttt{[10:03]} \textbf{B}: Okay.
            \item \texttt{[10:04]} \textbf{B}: The down one in mine is green.
            \item \texttt{[10:07]} \textbf{A}: Okay.
            \item \texttt{[10:08]} \textbf{A}: They are facing the right.
            \item \texttt{[10:10]} \textbf{B}: Correct.
            \item \texttt{[10:10]} \textbf{A}: Okay and they have
            \item \texttt{[10:12]} \textbf{A}: two fins.
            \item \texttt{[10:15]} \textbf{B}: Two.
            \item \texttt{[10:16]} \textbf{A}: Fins.
            \item \texttt{[10:16]} \textbf{B}: Yes.
            \item \texttt{[10:17]} \textbf{A}: aa so below the sharks I have three yellow
            \item \texttt{[10:21]} \textbf{A}: octopuses.
            \item \texttt{[10:23]} \textbf{B}: I have two.        
            \item ...
            \item \textcolor{violet}{\texttt{[19:02]} \textbf{A}: What were the colors of the animals right above the octopuses in my image?}
            \item \textcolor{purple}{\textbf{B}: There were blue and gray sharks above the octopuses in your image.}
            \item \textcolor{purple}{\textbf{Exact Match}: blue and gray}
        \end{dialoguelist}
    \end{dialogueboxB}
    \caption{Test example for spatial grounding from Spot the difference dataset}
    \label{fig:test_ex_1}
    \vspace{-15pt}
\end{figure}

\begin{figure}[htbp]
    \centering
    \begin{dialogueboxB}{Temporal} 
        \begin{dialoguelist}
            \item ...
            \item \texttt{[02:11]} \textbf{B}: Ahan.
            \item \texttt{[02:11]} \textbf{A}: \textless ahm\textgreater{}
            \item \texttt{[02:12]} \textbf{A}: To the left there is a cloud a gray cloud with raindrops underneath.
            \item \texttt{[02:18]} \textbf{B}: Oh so it's a white cloud.
            \item \texttt{[02:21]} \textbf{A}: Yeah.
            \item \texttt{[02:21]} \textbf{B}: And there's no rain drops.
            \item \texttt{[02:21]} \textbf{A}: Okay.
            \item ...
            \item \texttt{[10:45]} \textbf{A}: Let's switch to the next image now.
            \item ...
            \item \texttt{[12:04]} \textbf{A}: Okay and up in the sky to the right there's \textless aa\textgreater{} three clouds.
            \item \texttt{[12:09]} \textbf{B}: I have four.
            \item \texttt{[12:10]} \textbf{B}: \textless ahm\textgreater
            \item \texttt{[12:10]} \textbf{A}: Okay.
            \item \texttt{[12:11]} \textbf{B}: Two big ones on the right and two little ones \textless on the\textgreater{} on the left and upper side.
            \item \texttt{[12:17]} \textbf{A}: Okay I have three ones and 
            \item \texttt{[12:21]} \textbf{A}: I have \textless aa\textgreater{} the the small one is \textless aa\textgreater{} above the two big ones.
            \item \texttt{[12:27]} \textbf{B}: \textless ahm\textgreater
            \item \texttt{[12:28]} \textbf{B}: Okay \textless ahm\textgreater{}.
            \item ...
            \item \textcolor{violet}{\texttt{[14:07]} \textbf{B}: Do you remember the number of clouds in my last image?}
            \item \textcolor{purple}{\textbf{A}: You had four clouds.}
            \item \textcolor{purple}{\textbf{Exact Match}: four}
        \end{dialoguelist}
    \end{dialogueboxB}
    \caption{Test example for temporal grounding from Spot the difference dataset}
    \label{fig:test_ex_3}
\end{figure}

\section{LLM-As-A-Judge Effectiveness}
\label{apx:irr_llm}

To validate the reliability of our automated evaluator (LLM-as-a-Judge), we conducted a human evaluation study. We recruited 10 participants with a mixed background in Natural Language Processing, Cognitive Science and Neuro-Science. Each of them annotated a distinct set of 4 samples, resulting in a total evaluation set of $N=40$ instances. Participants were presented with a question, the gold reference answer, and the model-generated answer. They were instructed to determine whether the generated answer possessed the semantic equivalence of the gold answer. Treating the human annotations as ground truth, we calculated the accuracy of the LLM-Judge which achieved an accuracy of 90\% (36/40 matches with human judgment). It demonstrated a strong alignment between human and LLM judgements. Upon conducting a qualitative error analysis on the discrepancies, we observed that the LLM-Judge occasionally exhibits excessive leniency. It tends to overlook specific attribute mismatches if the core object class aligns. For example, given the question \textit{``What is on the table?''}, if the gold answer is \textit{blue bottle''} and the generated answer is \textit{``red bottle,''} the LLM may classify them as equivalent because both refer to a bottle, thereby ignoring the color hallucination. However, the high accuracy of LLMs as judges indicated their good overall performance that made us use them as automatic evaluators.

\section{Synthetic Data Generation}
\label{apx:pipeline}

We expand upon the Synthetic Data generation process to offer a comprehensive understanding of the framework's design and operation. The core objective of this generation process extends beyond mere factual accuracy to the creation of dialogs that are inherently spoken-like and situational. We specifically design the framework to model the complex dynamics of conversational grounding, where participants actively establish and maintain mutual understanding of entities and facts. This is achieved by programmatically incentivizing and modeling real-world conversational phenomena, including high volumes of back-and-forth exchanges, frequent backchanneling (e.g., "uh-huh," "yeah"), and explicit affirmations or clarifications of shared knowledge. By ensuring the agents are constantly listening to and reliably building upon each other's words, the resulting corpus aims to a deeper, more reliable form of collaborative communication.

\begin{figure*}[htbp]
  \centering
  \includegraphics[width=\linewidth, height=10.32cm]{
  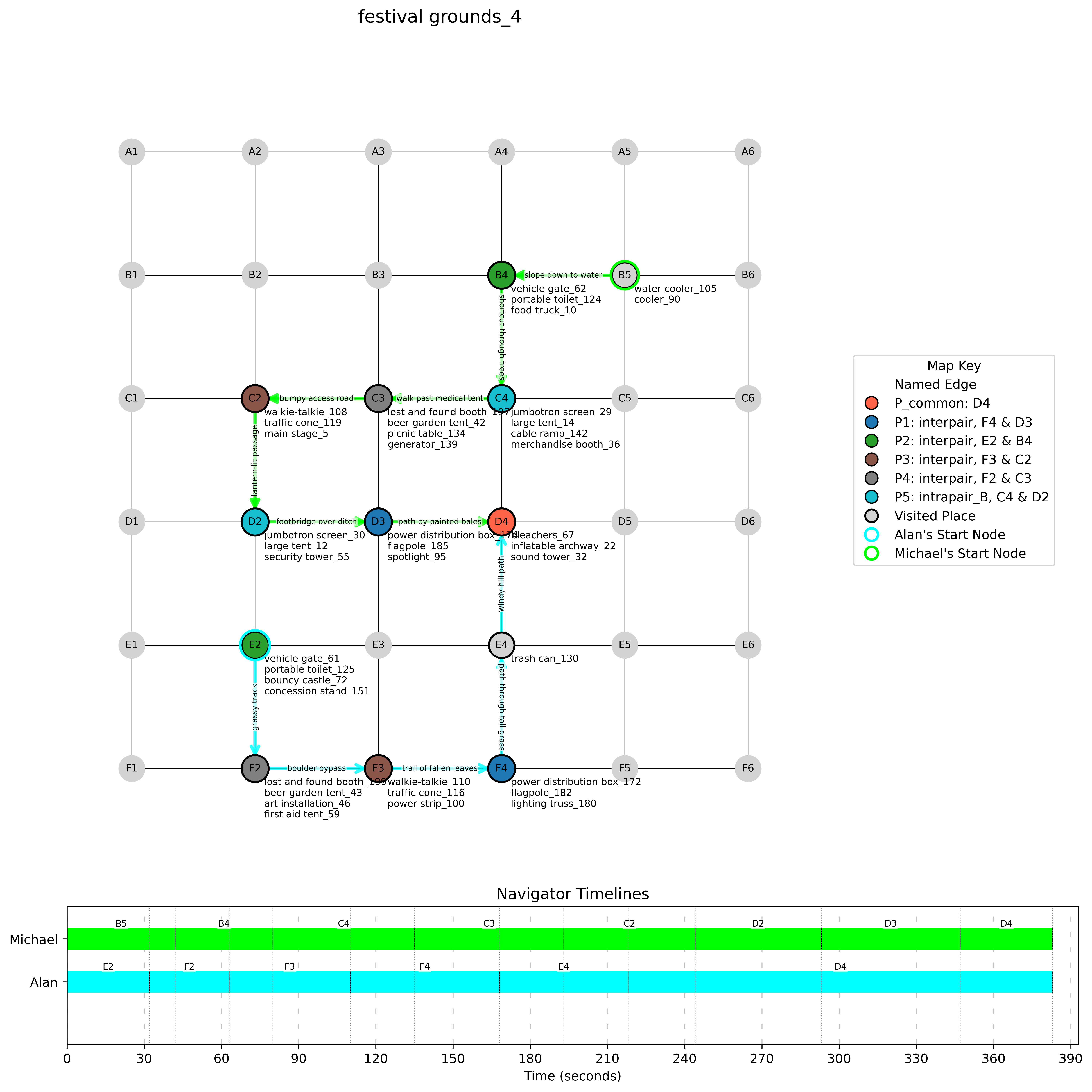}
\caption{A visualization of a \texttt{full\_scenario}. Unique objects, with their IDs, are crafted to have similar characteristics to induce ambiguity that has to be resolved. The ambiguity can arise from a navigator's experience alone (\texttt{intrapair}) or for their experience combined (\texttt{interpair}). Each chunks ($\mathcal{C}_{m,j}$) is delimited by the densely dotted lines.}
\label{fig:full_scene}
\vspace{-4mm}
\end{figure*}

\subsection{Grounded World Generation}

\textbf{The environment} is not a simple random collection of items; it's built from a structured taxonomy (see Figure \ref{fig:full_scene}). The process starts with a curated list of abstract places, like \texttt{festival grounds} or \texttt{historic museum}. For each place, we have generated inventories of objects and edges (paths). Objects are coherent with the place and highly detailed with properties and relations to other objects. This granular data provides the raw materials for a rich, fact-based world.

\textbf{The simulation} is a time-series event logger. Once a place is selected and populated with objects, two non-sentient virtual navigators are placed within it. The system then simulates their movement, creating a \texttt{full\_scenario} file that logs their `trajectories' over a set time span. This file is the definitive ground truth, recording what each navigator `sees', their paths, and their arrival/departure times. This detailed record ensures that every piece of information later referenced in the dialog is verifiable. The continuous Ground Truth is then partitioned into smaller chunks ($\mathcal{C}_{m,j}$), where $j$ denotes the navigator and $m$ denotes the index of the visited location $\mathcal{P}$ (or time segment). 

\subsection{Dialog Synthesis}

A programmatic "Script Writer" acts as the link between the simulated world and the LLM. It takes the structured information from a time segment, algorithmically ‘reasons’ over it, and converts this into a "script" prompt for the LLM to “perform” like an actor. This pre-processing is key to offloading the reasoning burden from the LLM. For example, the model will directly receive something like Listing \ref{lst:chunk}.

\begin{lstlisting}[
    language=json,
    caption=Example of a $\mathcal{C}_{m,j}$ chunk for the LLM,
    label={lst:chunk}
]
"place_type": "bakery production facility",
"time_span_s": [175, 212],
"navigator_A": {
  "name": "Michael",
  "zone_type": "packaging and storage zone",
  "status": "arrived & departing",
  "arrived_from": "office area connection",
  "objects_to_mention": [
    {"type": "packaging machine"},
    {
        "type": "label applicator",
        "property": "wipe-on",
        "relation": "holding",
        "reference": "label roll holder"
    },
    {"type": "delivery cart"},
    {"type": "refrigeration unit"}
  ]
}
... //continues
\end{lstlisting}

The Script Creator systematically introduces complexity to the dialog (i.e., noticing similarities, addressing ambiguities, etc.) This includes generating sentences that reference objects seen by the other navigator (indirect reference) or that require a deeper understanding of the shared context.

\begin{lstlisting}[
    language=json,
    caption=`Reasoned' part by the Script Writer,
    label={lst:chunk_solve}
]
... //continues
"current_observer": {
    "name": "Michael",
    "full_details_of_what_Michael_is_seeing": {
        "type": "delivery cart",
        "property": "fixed axle",
        "relation": "secured by",
        "reference": "securing strap"
    },
    "ask_question_about": "Michael asks about a delivery cart seen before by Kenneth",
    "to_avoid_in_question": "**Michael does NOT ask about specific properties that only Kenneth has seen.**"
},
"to_discuss_with": {
    "partner_name": "Kenneth",
    "full_details_of_what_Kenneth_saw": {
        "type": "delivery cart",
        "property": "solid rubber",
        "relation": "near",
        "reference": "loading ramp"
    },
    "seen_after_taking": "pallet movement lane",
    "that_place_also_had": "waste compactor"
}
\end{lstlisting}

The LLM then outputs the dialog:

\begin{lstlisting}[language=dialoguecsv,caption=Example of LLM-generated dialog segment, label={lst:dialogue_example_short}]
Michael:	I just arrived in the packaging and storage zone via the office area connection.
Kenneth:	Uh-huh, I'm still in the mixing and preparation zone.
Michael:	There's a packaging machine here with an adhesive seal protected by a safety interlock.
Kenneth:	Got it. My packaging machine here uses an ultrasonic seal and receives from a product feed conveyor.
Michael:	Noted. There's also a label applicator with a wipe-on method holding a label roll holder.
Kenneth:	Mine's a tamp-apply applicator triggered by a product sensor. Any delivery carts around?
Michael:	Yes, there's a delivery cart with a fixed axle secured by a securing strap.
Kenneth:	Mine's near a loading ramp with solid rubber wheels.
... //continues
\end{lstlisting}

\subsection{Q/A Synthesis}

First, we automatically identify "questionable moments" - where there are multiple entities of similar nature and a relational reference to them would only be resolved with a good grounding mechanism. We do this using both the ground truth ($\mathcal{C}_{\text{all}}$) and the dialog transcript ($\mathcal{D}$), algorithmically analyzing the conversation and flagging instances in a structured list.

A fact is preselected from the Ground Truth as the “expected answer”, and the LLM is prompted to generate a question that elicits this specific answer. The complex context for a moment is captured in a JSON fact:

\begin{lstlisting}[language=json,caption=Example of a fact for Q/A generation]
"fact_id": 17,
"chunk_id": "chunk_007",
"question_type": "inferred_comparison_result",
"questioner": "Kenneth",
"answerer": "Michael",
"context": {
    "property_of_obj_seen_by_Michael": "fixed axle",
    "property_of_obj_seen_by_Kenneth": "solid rubber",
    "ambiguous_object_type": "delivery cart",
    "disambiguating_fact": "the one that was where there was also a waste compactor"
},
"expected_answer": "solid rubber",
\end{lstlisting}

Given the whole context, the LLM generates the following questions in both a long and a short form giving us two variety of the same question:

\begin{lstlisting}[language=json,caption=Example of the generated and validated Question Pair]
"question_full": "Of the two delivery carts we discussed-yours with a fixed axle and mine with solid rubber wheels-which one was located where there was also a waste compactor?"
"question_short": "Which type of delivery cart was near the waste compactor?",
\end{lstlisting}

\subsection{Validation and Quality Control}

To ensure structural integrity and semantic accuracy, we employ an algorithmic validation pipeline active \textit{during generation}. The system parses the LLM output in real-time; if a check fails, a specific error message is fed back to the model to prompt self-correction. If the model fails repeatedly, the instance is discarded.
\textbf{Dialogs:} We enforce conversational pacing by limiting consecutive utterances, we verify that movement intentions are stated, and we ensure all mandatory objects seen in $\mathcal{C}_{m,j}$ are mentioned.
To account for linguistic variability, we employ a sliding-window semantic match (using \texttt{all-MiniLM-L6-v2} of the MiniLM family of models~\cite{minilm}) rather than strict keyword matching, verifying that specific properties and relations defined in the Ground Truth are contextually present.
\textbf{Questions:} We verify that generated questions contain mandatory scenario keywords, ensuring they are tied to the Ground Truth. Furthermore, we enforce strict constraints to prevent answer leakage and validate point-of-view consistency (e.g., ensuring correct pronoun usage for the observer vs. questioner).

Of all the questionable facts identified, only $\sim600$ ($\sim36\%$) of them were successfully converted into validated Q/A pairs. This high rejection rate indicates the effective pruning of logical inconsistencies and factual errors, significantly enhancing corpus quality.

\begin{table*}[htbp]
\centering
\resizebox{0.9\linewidth}{!}{%
\begin{tabular}{lcccccccc}
\toprule
\multirow{2}{*}{LLMs} & \multicolumn{4}{c}{Meetup} & \multicolumn{4}{c}{Spot the difference} \\ 
\cmidrule(lr){2-5} \cmidrule(lr){6-9}
                      & Temporal  & Spatial & Attributive & Inferred & Temporal & Spatial & Attributive & Inferred \\ \midrule
Full Dialog [from Table~\ref{Table:llm_result}]      &  0.20 / 0.18    &  0.18 / 0.16       &  0.24 / 0.26          &  0.26 / 0.16       &  0.18 / 0.12         & 0.18 / 0.10        &  0.20 / 0.12          &  0.16 / 0.10       \\ \midrule
Summarisation           &  0.16 / 0.14      &   0.12 / 0.12     &  0.20 / 0.14          &  0.18 / 0.14       &   0.12 / 0.10       &   0.14 / 0.12     &   0.12 / 0.10         &  0.08 / 0.02       \\ \midrule
Chunking (NV-Embbed-V2)         &  0.16 / 0.14         &  0.04 / 0       &   0.12 / 0.10         &  0.22 / 0.16       &    0 / 0      &   0.02 / 0      &  0.02 / 0          &  0.02 / 0       \\ \midrule
Chunking (BM25)          &  0.24 / 0.20         &  0.24 / 0.20      &   0.20 / 0.18        &  0.16 / 0.12      &  0 / 0       &  0 / 0       &  0.06 / 0.02        &  0.04 / 0      \\ \midrule
Agentic Ontology-based       &  0.40 / 0.36        &  0.32 / 0.26     &    0.26 / 0.22        & 0.20 / 0.16    &  0.12 / 0.06        &  0.10 / 0.04      &  0.14 / 0.08         & 0.06 / 0.10     \\ \bottomrule
\end{tabular}
}
\caption{FEM~($\uparrow$) / LLM-as-Judge~($\uparrow$) accuracy for various representations using Gemma2-2B. }
\label{Table:gemma2_2B_baseline}
\vspace{-1mm}
\end{table*}

\begin{table*}[htbp]
\centering
\resizebox{0.9\linewidth}{!}{%
\begin{tabular}{lcccccccc}
\toprule
\multirow{2}{*}{LLMs} & \multicolumn{4}{c}{Meetup} & \multicolumn{4}{c}{Spot the difference} \\ 
\cmidrule(lr){2-5} \cmidrule(lr){6-9}
                      & Temporal  & Spatial & Attributive & Inferred & Temporal & Spatial & Attributive & Inferred \\ \midrule
Full Dialog [from Table~\ref{Table:llm_result}]      &  0.48 / 0.38         &   0.50 / 0.46     &  0.38 / 0.40          &  0.26 / 0.18       & 0.34 / 0.20        &  0.24 / 0.16       & 0.28 / 0.16           & 0.20 / 0.10        \\ \midrule
Summarisation           & 0.28 / 0.22       &  0.24 / 0.18      &  0.28 / 0.24          &   0.16 / 0.14      &   0.12 / 0.10       &   0.10 / 0.10     &    0.12 / 0.12        &    0.16 / 0.14     \\ \midrule
Chunking (NV-Embbed-V2)         &  0.22 / 0.16          &   0.10 / 0.10      & 0.18 / 0.16           &   0.22 / 0.22      & 0 / 0         &  0 / 0.02       &   0 / 0         &  0.04 / 0       \\ \midrule
Chunking (BM25)          & 0.34 / 0.26          &  0.22 / 0.24      &   0.24 / 0.26        &  0.26 / 0.24      &   0.02 / 0      &  0.02 / 0       &  0.02 / 0        & 0.06 / 0.02       \\ \midrule
Agentic Ontology-based       &  0.44 / 0.46         &  0.40 / 0.38     &   0.42 / 0.36       &  0.26 / 0.24     &  0.32 / 0.32        &   0.36 / 0.32     &   0.30 / 0.26        & 0.28 / 0.16     \\ \bottomrule
\end{tabular}
}
\caption{FEM~($\uparrow$) / LLM-as-Judge~($\uparrow$) accuracy for various representations using Gemma2-9B. }
\label{Table:gemma2_9B_baseline}
\vspace{-1mm}
\end{table*}

\section{GRPO Training}
\label{apx:grpo}

\begin{figure}[ht]
  \includegraphics[width=0.9\columnwidth]{
  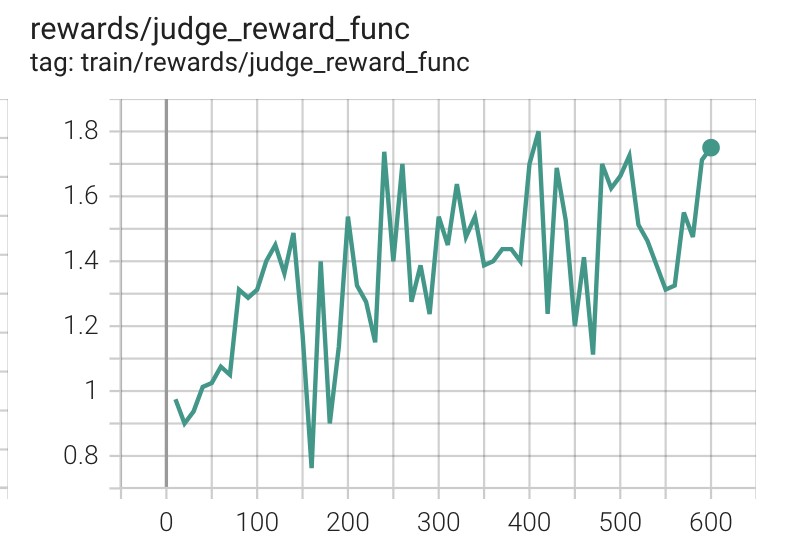}
\caption{Training reward over time while using the synthetic data. Here the y-axis is the reward while x-axis is the step.}
\label{fig:rl}
\vspace{-4mm}
\end{figure}

For our GRPO training, we used 600 Question/Answer pairs from 60 artificially generated dialogs. While we can generate much more synthetic data, we were contrained by the computing resources and decided to use only 600 instances which showed to be enough for the model as the reward stabilized towards the end of the training as seen in Figure~\ref{fig:rl}. The huggingface TRL~\cite{trl} library was used for GRPO training. 

\subsection{Hyperparameters}

We used a batch size of 16 for the RL training. The training used 8 A100 GPUs in total out of which 4 A100s were used by VLLM for the inference. Here are our other hyperparameters : learning rate - 1e-6, GRPO beta - 0.025, weight\_decay - 0.1, adam\_beta1 - 0.9, adam\_beta2 - 0.99, warm-up ratio - 0.1, and a cosine learning scheduler.

\section{Gemma Results}
\label{apx:test_result}

Tables~\ref{Table:gemma2_2B_baseline} and \ref{Table:gemma2_9B_baseline} contain all the results for Gemma2 2B and 9B models. As seen from the Tables, the overall pattern remains the same as compared to Llama 3.1-8B.
The 2 Billion model performs relatively poor compared to the 9 Billion model across all the categories. Chunking in general provides the worst results among all the representation techniques suggesting the importance of good abstraction of information that preserves the links between the information across different segments of the conversation. Surprisingly, for the gemma2 models that struggled with longer context of STD, the agentic system performs better than when provided with the full dialog for Meetup test cases. This suggests that providing a structure might help it reason better for longer contexts.

\section{Prompts}
\label{apx:prompts}


During testing, we utilized the system prompts shown in Figures~\ref{fig:judge_prompt}, \ref{fig:full_prompt} and \ref{fig:in_context_prompt}. Our approach leverages chain-of-thought prompting (Figure~\ref{fig:full_prompt}), which we found to be more effective than direct questioning. This prompt outlines the game's rules to help the LLM understand the context. However, it differs from the in-context prompt (Figure~\ref{fig:in_context_prompt}) by not specifying how to formulate the answer. We have omitted the prompts for our Agentic method from this paper due to the numerous large prompts used in its different phases. Furthermore, we deliberately excluded textual descriptions of the images from the prompts. This ensures that the questions can only be answered using the conversational context, thereby testing the model's ability to form an accurate common ground representation based on what has been discussed.

\section{License}
All tools and models used in this research were employed under licenses that permit academic use. Specifically, we utilized the Huggingface TRL library and the Sentence Transformer model all-MiniLM-L6-v2 under the Apache-2.0 license. The Llama 3.1, Gemma 2, and Qwen models were used in accordance with the Llama 3.1 Community License Agreement, Google's usage license, and the Apache-2.0 license respectively.

\section{Use of AI}
We declare that we have used LLMs for helping us paraphrase some of the sentences in this paper in order to better convey our findngs to the community. However, we would like to mention that it was not used for writing big chunks of the paper. Every single line written here has been thoroughly read and revised by the authors.

\begin{figure*}[t]
    \centering 
    \begin{dialogueboxC}{Judge prompt}
    You are a strict evaluator. Given the Question, LLM response and the correct response, judge whether the LLM response and the correct response both have the same meaning provided the question. \\

    *KEEP IN MIND TO ALWAYS FOLLOW THESE RULES* - \\
    
    1) You have to give 'DIFFERENT' if they are not having the same meaning and 'SAME' if they mean the same. \\
    2) If the words are synonyms then they shall be considered the same.\\ 
    3) If the response is a negation of the other response then it does not have the same meaning and hence should be 'DIFFERENT'. \\
    4) If correct answer is kitchen but the response is dining room, then it should be 'DIFFERENT' because they are different. Do not reason that they have similar purpose. Just because the place might have similar function does not make it the same.\\
    5) Sometimes the answers from LLM might contain reasoning, but you should only provide the score based on the final answer provided which is at the end of the answer.\\
    6) If the two responses are negations but in different styles, they should still be considered as the 'SAME'. eg. 'do not recall presence of a cat' is the same as 'no cats were there' are the same, the first one is more indirect form of saying it.\\
    7) If the main content of the answers are the same then answer 'SAME'. The style might be different. For example, 'yes there was a car there' and 'a car was there' are the 'SAME' in meaning even though one is more affirmative than the other.\\

    You have to give the answer in the following format:\\
    \begin{verbatim}
    <reasoning>
    (your reasoning here in one or two sentences where you concretely mention
    the reason why they mean the same or not.)
    </reasoning>
    <answer>
    (your final verdict here)
    </answer>
    \end{verbatim}
    \end{dialogueboxC}

    \caption{The system prompt for LLM as a judge.}
    \label{fig:judge_prompt}
\end{figure*}

\begin{figure*}[t]
    \centering 
    \begin{dialogueboxC}{Full dialog prompt}
        You are participant \{current\_participant\} who is navigating a shared environment with another participant \{other\_participant\}. You are in a virtual space that has lots of different areas. The participants are initially located in different areas. \\
        
        The goal is for the two participants to locate themselves in the same area. In order to achieve this goal, the participants communicate with one another and describe the area they find themselves in. On the basis of those descriptions, they move to different areas and describe their new areas to the other participant. \\
        
        The two participants need to find themselves in the same area. In the last utterance, participant \{other\_participant\} asks a question. Think about the question, get all the information required from the dialog as participant \{current\_participant\} and then answer that question as your response. First reason and interpret the question. Identify whose experience you have to refer to \{other\_participant\} or \{current\_participant\} or both. Fully break it down and then give the answer. Reason step by step and then answer. \\
        
        The final answer should be in the format - 
        \begin{verbatim}
            <reasoning> 
            (your reasoning here which requires analysing the dialog and understanding 
            what are the important information to look at before answering. Take all the 
            important information into consideration step by step in your reasoning.) 
            </reasoning> 
            <answer> 
            (your final answer here. DO NOT REASON HERE!!) 
            </answer>.
        \end{verbatim}
    \end{dialogueboxC}

    \caption{The system prompt for the LLM when provided with the full dialog for Meetup test cases. It was also used for the GRPO trained models.}
    \label{fig:full_prompt}
\end{figure*}

\begin{figure*}[t]
    \centering 
    \begin{dialogueboxC}{In-Context Prompt}
    [FIRST PARAGRAPH SAME AS Figure~\ref{fig:full_prompt}] \\
        
    The two participants need to find themselves in the same area. In the last utterance, participant \{other\_participant\} asks a question.
    Answer the question. First understand what it means and then answer as the other user. In other words, if A asked the question then answer as if you are B and vice versa. Remember they start at different locations and try and meet and are may be in different rooms and may have parallel observations which are valid from their perspective. Identify whose experience and observations you have to refer to. First reason and interpret the question. Identify whose experince you have to refer to, then A or B or both fully break it down and then give the answer. \\
        
    Remember most importantly -\\
        
    1) Reason step by step and then answer. 
    Example:
        
    Question from A: do you remember the park I visited?
        
    Although B has to answer here, the reference is to A's experience, so check what A observed.
        
    Similarly, For, B: Do you remember the garden i visited?
    -> A should answer based on what B experienced. \\
        
    2) Most Crucial: If the question is about "my x" (e.g., "my dining room", "my office"), always refer to the one who asked the question. 
    So:
        
    - "My dining room" (asked by A) → Refer to A's observation of their dining room.\\
    - "My chandelier" (asked by B) → Refer to B's observation of their chandelier. \\
        
    3) Also, prioritize semantic ownership and pronoun resolution carefully. Don't let "described by B" override the ownership stated in the question.
    Always prioritize the experience of the person who owns the object/location.
    If A says "My chandelier," then even if B is answering, the correct grounding is in A's description — because A is the owner. Use the other speaker's description if that's whose experience is referenced in the question. \\

    4) If observations conflict strongly, assume two different entities till it has been resolved. Example conflict:
    A: "I see one white fridge", B: "I have five yellow fridges"
    → this means they are in two different kitchens. \\

    Example grounding from dialog:\\
    00:02:01 — A: I’m in a kitchen and see a fridge\\
    00:02:10 — B:  yellow?\\
    00:02:15 — A:   No not yellow but white.\\
    00:03:10 - A: Found another kitchen. White walls.\\
    Spatial Grounding: A is in Kitchen A, B is in Kitchen B. A then moves to kitchen C.\\
    Object Grounding: Fridge A (white) in Kitchen A — seen by A, Fridge B (yellow) in Kitchen B — seen by B, Kitchen C has white walls - seen by A. \\
    Temporal Grounding: A first goes to kitchen A, B simultaniously goes to kitchen B. A then move and finds another kitchen C which is the second kitchen that A saw. \\
    
    The final answer should be in the format -
        
    \begin{verbatim}
    <reasoning> ... [Like in Figure~\ref{fig:full_prompt}] ... </answer>
    \end{verbatim}
    \end{dialogueboxC}

    \caption{The system prompt for in-context learning with full dialog history for Meetup test cases.}
    \label{fig:in_context_prompt}
\end{figure*}

\begin{figure*}[t]
    \centering 
    \begin{dialogueboxD}{Qwen QWQ Failure}
        \textbf{Question} : Can you please remind me which room is positioned east of the dining room that I visited? \\

        \textbf{Correct Answer} : Attic \\

        \textbf{Generated Answer} : Okay, let's see what the user needs here. They're asking for the next utterance in the conversation based on the provided dialog history. The last message from A is a question: \"Can you remind me which room is positioned east of the dining room that I visited?\" So, I need to figure out how B would respond.
        
        Looking through the conversation history, we need to track where each participant has been. Let's start by recalling the key points: 
        First, Participant B started in a dining room with a red chandelier (timestamp [00:00:27]). Then, A moved to the east and mentioned being in an attic ([00:01:04]), but then later described being in a room with a curved ceiling, ceiling fan, and triangular shape, possibly an attic again. 
        At [00:02:43], A mentions finding a jacuzzi with beach chairs, which seems like a bathroom or maybe a spa room. But then moves to a pool area ([00:03:31] mentions brown tiles and pool). Meanwhile, B went to a street ([00:03:48]) and then an attic with a bed and closet, which was different from A's attic without a bed. Eventually, B ended up in the room A was referring to, which had a window and a wooden table on the right, plus a 2007 date ([00:05:04]).
        The question now is about the room east of the original dining room. From earlier messages, when A first said they were in the dining room with a red chandelier, then moved east to what they thought was an attic. However, after moving, the next image B received was a library/museum building ([00:01:16 Image B]). Wait, actually, the timestamps show that after A says they're heading east, the next Image B is the library building. Hmm, but maybe the movement directions correspond to changing rooms. Since in virtual spaces, moving east could lead them to adjacent rooms. 
        Wait, the initial dining room was B's starting point. When A was in the dining room too? At [00:01:04], A says \"in dining room with red chandelier,\" so they were both in the same dining room at some point. But since they needed to get to the same room, perhaps moving east from there led A to a different room. The Image B after A's move to east showed the library/museum building. So maybe moving east from the dining room leads to the library? Because the next image after A moving east was the library. So when A asks which room is east of the dining room, B would refer to that library or museum.
        But looking at the latest interactions, B is currently in a room with a window, wooden table on the right, and a 2007 date. That room's description ([00:05:04]) is an empty room with a dining table, chairs, desk, laptop, etc. Maybe that's a different area. Alternatively, B might remember previous movements. Alternatively, maybe the library was the room east of dining because after A went east, the next image was the library. Therefore, B would say something like \"the library\" or \"museum building\". Also, considering that in the final messages, they ended up in a room with the 2007 date, which might not be the library, but another spot. So the response should be B telling A that the room east of the dining room is the library or museum. Let me check the exact description again. The Image B after A moved east was: \"large, unique building with a distinctive dome...library or museum\". So B should mention that. Hence, the answer is probably pointing out the library/museum as the eastern room.</think>
        
        [00:05:42] B: The room east of the dining room looked like a library or museum with a big dome. check the building with cars parked out front and traffic lights
        
    \end{dialogueboxD}

    \caption{Failure case from Qwen-QWQ where it produces a long answer despite reasoning in it's <reasoning> tags separately}
    \label{fig:Qwen_wrong}
\end{figure*}

%% file: acl_latex.bbl
\begin{thebibliography}{50}
\providecommand{\natexlab}[1]{#1}

\bibitem[{Allison et~al.(2006)Allison, Guthrie, and Guthrie}]{sparsity}
Ben Allison, David Guthrie, and Louise Guthrie. 2006.
\newblock Another look at the data sparsity problem.
\newblock In \emph{Text, Speech and Dialogue}, pages 327--334, Berlin, Heidelberg. Springer Berlin Heidelberg.

\bibitem[{Blache and Houlès(2021)}]{frames}
Philippe Blache and Matthis Houlès. 2021.
\newblock \href {https://doi.org/10.5121/csit.2021.112003} {Common ground, frames and slots for comprehension in dialogue systems}.
\newblock In \emph{7th International Conference on Natural Language Computing}, pages 33--43.

\bibitem[{Bu et~al.(2025)Bu, Chang, Chen, Dang, Wu, He, and Wu}]{bu-etal-2025-query}
Chenyang Bu, Guojie Chang, Zihao Chen, CunYuan Dang, Zhize Wu, Yi~He, and Xindong Wu. 2025.
\newblock \href {https://doi.org/10.18653/v1/2025.findings-acl.1100} {Query-driven multimodal {G}raph{RAG}: Dynamic local knowledge graph construction for online reasoning}.
\newblock In \emph{Findings of the Association for Computational Linguistics: ACL 2025}, pages 21360--21380, Vienna, Austria. Association for Computational Linguistics.

\bibitem[{Cassell et~al.(2001)Cassell, Bickmore, Campbell, Vilhjálmsson, and Yan}]{CASSELL200155}
J~Cassell, T~Bickmore, L~Campbell, H~Vilhjálmsson, and H~Yan. 2001.
\newblock \href {https://doi.org/10.1016/S0950-7051(00)00102-7} {More than just a pretty face: conversational protocols and the affordances of embodiment}.
\newblock \emph{Knowledge-Based Systems}, 14(1):55--64.

\bibitem[{Chase(2022)}]{langchain}
Harrison Chase. 2022.
\newblock Langchain.
\newblock \url{https://github.com/langchain-ai/langchain}.

\bibitem[{Chen et~al.(2025)Chen, Kim, Patterson, Breazeal, and Park}]{soc_rob}
Huili Chen, Yubin Kim, Kejia Patterson, Cynthia Breazeal, and Hae~Won Park. 2025.
\newblock \href {https://doi.org/10.1126/scirobotics.adk3307} {Social robots as conversational catalysts: Enhancing long-term human-human interaction at home}.
\newblock \emph{Science Robotics}, 10(100):eadk3307.

\bibitem[{Clark and Schaefer(1989)}]{Clark1989ContributingTD}
Herbert~H. Clark and Ed~Schaefer. 1989.
\newblock Contributing to discourse.
\newblock \emph{Cogn. Sci.}, 13:259--294.

\bibitem[{Crible(2022)}]{phenomena}
Ludivine Crible. 2022.
\newblock \href {https://doi.org/10.4000/discours.12024} {Studying discourse from corpus and experimental data: Bridging the methodological gap}.
\newblock \emph{Discours}, 30.

\bibitem[{Denis(2010)}]{DRT}
Alexandre Denis. 2010.
\newblock \href {https://aclanthology.org/W10-4203} {Generating referring expressions with reference domain theory}.
\newblock In \emph{Proceedings of the 6th International Natural Language Generation Conference}. Association for Computational Linguistics.

\bibitem[{Di~Maro et~al.(2021)Di~Maro, Origlia, and Cutugno}]{butter}
Maria Di~Maro, Antonio Origlia, and Francesco Cutugno. 2021.
\newblock Cutting melted butter? common ground inconsistencies management in dialogue systems using graph databases.
\newblock In \emph{IJCoL, 7-1, 2}, pages 157--190.

\bibitem[{Dong et~al.(2022)Dong, Li, Dai, Zheng, Wu, Chang, Sun, Xu, Li, and Sui}]{incontext}
Qingxiu Dong, Lei Li, Damai Dai, Ce~Zheng, Zhiyong Wu, Baobao Chang, Xu~Sun, Jingjing Xu, Lei Li, and Zhifang Sui. 2022.
\newblock \href {https://api.semanticscholar.org/CorpusID:255372865} {A survey on in-context learning}.
\newblock In \emph{Conference on Empirical Methods in Natural Language Processing}.

\bibitem[{Gemma(2024)}]{gemma2}
Team Gemma. 2024.
\newblock \href {https://api.semanticscholar.org/CorpusID:270843326} {Gemma 2: Improving open language models at a practical size}.
\newblock \emph{ArXiv}, abs/2408.00118.

\bibitem[{Henaff et~al.(2016)Henaff, Weston, Szlam, Bordes, and LeCun}]{Henaff2016TrackingTW}
Mikael Henaff, Jason Weston, Arthur Szlam, Antoine Bordes, and Yann LeCun. 2016.
\newblock \href {https://api.semanticscholar.org/CorpusID:11243593} {Tracking the world state with recurrent entity networks}.
\newblock \emph{ArXiv}, abs/1612.03969.

\bibitem[{Ilinykh et~al.(2019)Ilinykh, Zarrie{\ss}, and Schlangen}]{meetup}
Nikolai Ilinykh, Sina Zarrie{\ss}, and David Schlangen. 2019.
\newblock \href {http://semdial.org/anthology/Z19-Ilinykh_semdial_0006.pdf} {Meet up! a corpus of joint activity dialogues in a visual environment}.
\newblock In \emph{Proceedings of the 23rd Workshop on the Semantics and Pragmatics of Dialogue - Full Papers}, London, United Kingdom. SEMDIAL.

\bibitem[{Jacqmin et~al.(2022)Jacqmin, Rojas~Barahona, and Favre}]{jacqmin-etal-2022-follow}
L{\'e}o Jacqmin, Lina~M. Rojas~Barahona, and Benoit Favre. 2022.
\newblock \href {https://doi.org/10.18653/v1/2022.sigdial-1.33} {``do you follow me?'': A survey of recent approaches in dialogue state tracking}.
\newblock In \emph{Proceedings of the 23rd Annual Meeting of the Special Interest Group on Discourse and Dialogue}, pages 336--350, Edinburgh, UK. Association for Computational Linguistics.

\bibitem[{Khebour et~al.(2024)Khebour, Lai, Bradford, Zhu, Brutti, Tam, Tu, Ibarra, Blanchard, Krishnaswamy, and Pustejovsky}]{khebour-etal-2024-common}
Ibrahim~Khalil Khebour, Kenneth Lai, Mariah Bradford, Yifan Zhu, Richard~A. Brutti, Christopher Tam, Jingxuan Tu, Benjamin~A. Ibarra, Nathaniel Blanchard, Nikhil Krishnaswamy, and James Pustejovsky. 2024.
\newblock \href {https://aclanthology.org/2024.lrec-main.318/} {Common ground tracking in multimodal dialogue}.
\newblock In \emph{Proceedings of the 2024 Joint International Conference on Computational Linguistics, Language Resources and Evaluation (LREC-COLING 2024)}, pages 3587--3602, Torino, Italia. ELRA and ICCL.

\bibitem[{Kruijt and Vossen(2022)}]{kruijt-vossen-2022-role}
Jaap Kruijt and Piek Vossen. 2022.
\newblock \href {https://aclanthology.org/2022.crac-1.10/} {The role of common ground for referential expressions in social dialogues}.
\newblock In \emph{Proceedings of the Fifth Workshop on Computational Models of Reference, Anaphora and Coreference}, pages 99--110, Gyeongju, Republic of Korea. Association for Computational Linguistics.

\bibitem[{Lee et~al.(2024)Lee, Roy, Xu, Raiman, Shoeybi, Catanzaro, and Ping}]{lee2024nv}
Chankyu Lee, Rajarshi Roy, Mengyao Xu, Jonathan Raiman, Mohammad Shoeybi, Bryan Catanzaro, and Wei Ping. 2024.
\newblock Nv-embed: Improved techniques for training llms as generalist embedding models.
\newblock \emph{arXiv preprint arXiv:2405.17428}.

\bibitem[{Lewis et~al.(2020)Lewis, Perez, Piktus, Petroni, Karpukhin, Goyal, K\"{u}ttler, Lewis, Yih, Rockt\"{a}schel, Riedel, and Kiela}]{lewis-et-al}
Patrick Lewis, Ethan Perez, Aleksandra Piktus, Fabio Petroni, Vladimir Karpukhin, Naman Goyal, Heinrich K\"{u}ttler, Mike Lewis, Wen-tau Yih, Tim Rockt\"{a}schel, Sebastian Riedel, and Douwe Kiela. 2020.
\newblock Retrieval-augmented generation for knowledge-intensive nlp tasks.
\newblock In \emph{Proceedings of the 34th International Conference on Neural Information Processing Systems}, NIPS '20, Red Hook, NY, USA. Curran Associates Inc.

\bibitem[{Llama(2024)}]{llama3}
Team Llama. 2024.
\newblock \href {https://api.semanticscholar.org/CorpusID:271571434} {The llama 3 herd of models}.
\newblock \emph{ArXiv}, abs/2407.21783.

\bibitem[{Lopes et~al.(2018)Lopes, Hemmingsson, and {\AA}strand}]{spotthedifference}
Jos{\'e} Lopes, Nils Hemmingsson, and Oliver {\AA}strand. 2018.
\newblock \href {https://aclanthology.org/L18-1305} {The spot the difference corpus: a multi-modal corpus of spontaneous task oriented spoken interactions}.
\newblock In \emph{Proceedings of the Eleventh International Conference on Language Resources and Evaluation ({LREC} 2018)}, Miyazaki, Japan. European Language Resources Association (ELRA).

\bibitem[{Madge et~al.(2025)Madge, Purver, and Poesio}]{Madge2025ReferentialAA}
Chris Madge, Matthew Purver, and Massimo Poesio. 2025.
\newblock \href {https://api.semanticscholar.org/CorpusID:280276495} {Referential ambiguity and clarification requests: comparing human and llm behaviour}.
\newblock \emph{ArXiv}, abs/2507.10445.

\bibitem[{Mohapatra et~al.(2024{\natexlab{a}})Mohapatra, Hassan, Romary, and Cassell}]{Mohapatra2024ConversationalGA}
Biswesh Mohapatra, Seemab Hassan, Laurent Romary, and Justine Cassell. 2024{\natexlab{a}}.
\newblock \href {https://api.semanticscholar.org/CorpusID:268680732} {Conversational grounding: Annotation and analysis of grounding acts and grounding units}.
\newblock In \emph{Proceedings of LREC-COLING}. Association for Computational Linguistics.

\bibitem[{Mohapatra et~al.(2024{\natexlab{b}})Mohapatra, Kapadnis, Romary, and Cassell}]{Mohapatra2024EvaluatingTE}
Biswesh Mohapatra, Manav~Nitin Kapadnis, Laurent Romary, and Justine Cassell. 2024{\natexlab{b}}.
\newblock \href {https://api.semanticscholar.org/CorpusID:273901650} {Evaluating the effectiveness of large language models in establishing conversational grounding}.
\newblock In \emph{Conference on Empirical Methods in Natural Language Processing}.

\bibitem[{Mohapatra et~al.(2021)Mohapatra, Pandey, Contractor, and Joshi}]{simulated-chats}
Biswesh Mohapatra, Gaurav Pandey, Danish Contractor, and Sachindra Joshi. 2021.
\newblock \href {https://doi.org/10.18653/v1/2021.findings-emnlp.103} {Simulated chats for building dialog systems: Learning to generate conversations from instructions}.
\newblock In \emph{Findings of the Association for Computational Linguistics: EMNLP 2021}, pages 1190--1203, Punta Cana, Dominican Republic. Association for Computational Linguistics.

\bibitem[{Padmakumar et~al.(2023)Padmakumar, Hedayatnia, Jin, Lange, Kim, Peng, Liu, and Hakkani-Tur}]{padmakumar-etal-2023-investigating}
Vishakh Padmakumar, Behnam Hedayatnia, Di~Jin, Patrick Lange, Seokhwan Kim, Nanyun Peng, Yang Liu, and Dilek Hakkani-Tur. 2023.
\newblock \href {https://doi.org/10.18653/v1/2023.sigdial-1.50} {Investigating the representation of open domain dialogue context for transformer models}.
\newblock In \emph{Proceedings of the 24th Annual Meeting of the Special Interest Group on Discourse and Dialogue}, pages 538--547, Prague, Czechia. Association for Computational Linguistics.

\bibitem[{Qiu et~al.(2024)Qiu, Liu, Li, Zhu, and Zheng}]{minddial}
Shuwen Qiu, Mingdian Liu, Hengli Li, Song-Chun Zhu, and Zilong Zheng. 2024.
\newblock \href {https://doi.org/10.18653/v1/2024.sigdial-1.63} {{M}ind{D}ial: Enhancing conversational agents with theory-of-mind for common ground alignment and negotiation}.
\newblock In \emph{Proceedings of the 25th Annual Meeting of the Special Interest Group on Discourse and Dialogue}, pages 746--759, Kyoto, Japan. Association for Computational Linguistics.

\bibitem[{Qwen(2025)}]{qwenqwq}
Team Qwen. 2025.
\newblock \href {https://qwenlm.github.io/blog/qwq-32b/} {Qwq-32b: Embracing the power of reinforcement learning}.

\bibitem[{Rajendran et~al.(2019)Rajendran, Ganhotra, and Polymenakos}]{rajendran-etal-2019-learning}
Janarthanan Rajendran, Jatin Ganhotra, and Lazaros~C. Polymenakos. 2019.
\newblock \href {https://doi.org/10.1162/tacl_a_00274} {Learning end-to-end goal-oriented dialog with maximal user task success and minimal human agent use}.
\newblock \emph{Transactions of the Association for Computational Linguistics}, 7:375--386.

\bibitem[{Ramprasad et~al.(2024)Ramprasad, Ferracane, and Lipton}]{Ramprasad2024AnalyzingLB}
Sanjana Ramprasad, Elisa Ferracane, and Zachary~Chase Lipton. 2024.
\newblock \href {https://api.semanticscholar.org/CorpusID:270257966} {Analyzing llm behavior in dialogue summarization: Unveiling circumstantial hallucination trends}.
\newblock In \emph{Annual Meeting of the Association for Computational Linguistics}.

\bibitem[{Robertson et~al.(1994)Robertson, Walker, Jones, Hancock-Beaulieu, and Gatford}]{bm25}
Stephen Robertson, Steve Walker, Susan Jones, Micheline Hancock-Beaulieu, and Mike Gatford. 1994.
\newblock Okapi at trec-3.
\newblock In \emph{Text Retrieval Conference}, pages 0--.

\bibitem[{Schlangen and Skantze(2009)}]{schlangen-skantze-2009-general}
David Schlangen and Gabriel Skantze. 2009.
\newblock \href {https://aclanthology.org/E09-1081/} {A general, abstract model of incremental dialogue processing}.
\newblock In \emph{Proceedings of the 12th Conference of the {E}uropean Chapter of the {ACL} ({EACL} 2009)}, pages 710--718, Athens, Greece. Association for Computational Linguistics.

\bibitem[{Shao et~al.(2024)Shao, Wang, Zhu, Xu, Song, Zhang, Li, Wu, and Guo}]{grpo}
Zhihong Shao, Peiyi Wang, Qihao Zhu, Runxin Xu, Jun-Mei Song, Mingchuan Zhang, Y.~K. Li, Yu~Wu, and Daya Guo. 2024.
\newblock \href {https://api.semanticscholar.org/CorpusID:267412607} {Deepseekmath: Pushing the limits of mathematical reasoning in open language models}.
\newblock \emph{ArXiv}, abs/2402.03300.

\bibitem[{Shopen(1985)}]{diectic}
T.~Shopen. 1985.
\newblock \href {https://books.google.fr/books?id=iz4HQmlj6v8C} {\emph{Language Typology and Syntactic Description: Volume 3}}.
\newblock Language Typology and Syntactic Description. Cambridge University Press.

\bibitem[{Shridhar et~al.(2020)Shridhar, Mittal, and Hsu}]{ingress}
Mohit Shridhar, Dixant Mittal, and David Hsu. 2020.
\newblock \href {https://doi.org/10.1177/0278364919897133} {Ingress: Interactive visual grounding of referring expressions}.
\newblock \emph{Int. J. Rob. Res.}, 39(2–3):217–232.

\bibitem[{Shuster et~al.(2021)Shuster, Poff, Chen, Kiela, and Weston}]{shuster-etal-2021}
Kurt Shuster, Spencer Poff, Moya Chen, Douwe Kiela, and Jason Weston. 2021.
\newblock \href {https://doi.org/10.18653/v1/2021.findings-emnlp.320} {Retrieval augmentation reduces hallucination in conversation}.
\newblock In \emph{Findings of the Association for Computational Linguistics: EMNLP 2021}, pages 3784--3803, Punta Cana, Dominican Republic. Association for Computational Linguistics.

\bibitem[{Smith(2015)}]{smith2015analog}
Gillian Smith. 2015.
\newblock An analog history of procedural content generation.
\newblock In \emph{FDG}, pages 1--6. Boston, MA.

\bibitem[{Suresh et~al.(2025)Suresh, Mengjun, Pranav, and Chng}]{suresh2024diasynth}
Sathya~Krishnan Suresh, Wu~Mengjun, Tushar Pranav, and EngSiong Chng. 2025.
\newblock \href {https://doi.org/10.18653/v1/2025.findings-naacl.40} {{D}ia{S}ynth: Synthetic dialogue generation framework for low resource dialogue applications}.
\newblock In \emph{Findings of the Association for Computational Linguistics: NAACL 2025}, pages 673--690, Albuquerque, New Mexico. Association for Computational Linguistics.

\bibitem[{Traum and Allen(1994)}]{TraumGA}
David Traum and James Allen. 1994.
\newblock A "speech acts" approach to grounding in conversation.
\newblock \emph{Proceedings of International Conference on Spoken Language Processing}.

\bibitem[{Traum and Larsson(2003)}]{Traum2003}
David~R. Traum and Staffan Larsson. 2003.
\newblock \href {https://doi.org/10.1007/978-94-010-0019-2_15} {\emph{The Information State Approach to Dialogue Management}}, pages 325--353.
\newblock Springer Netherlands, Dordrecht.

\bibitem[{von Werra et~al.(2020)von Werra, Belkada, Tunstall, Beeching, Thrush, Lambert, Huang, Rasul, and Gallouédec}]{trl}
Leandro von Werra, Younes Belkada, Lewis Tunstall, Edward Beeching, Tristan Thrush, Nathan Lambert, Shengyi Huang, Kashif Rasul, and Quentin Gallouédec. 2020.
\newblock Trl: Transformer reinforcement learning.
\newblock \url{https://github.com/huggingface/trl}.

\bibitem[{Walker et~al.(2022)Walker, Ultes, and Lison}]{graphWOZ}
Nicholas~Thomas Walker, Stefan Ultes, and Pierre Lison. 2022.
\newblock \href {https://api.semanticscholar.org/CorpusID:253801560} {Graphwoz: Dialogue management with conversational knowledge graphs}.
\newblock \emph{ArXiv}, abs/2211.12852.

\bibitem[{Wang et~al.(2020)Wang, Wei, Dong, Bao, Yang, and Zhou}]{minilm}
Wenhui Wang, Furu Wei, Li~Dong, Hangbo Bao, Nan Yang, and Ming Zhou. 2020.
\newblock Minilm: deep self-attention distillation for task-agnostic compression of pre-trained transformers.
\newblock In \emph{Proceedings of the 34th International Conference on Neural Information Processing Systems}, NIPS '20, Red Hook, NY, USA. Curran Associates Inc.

\bibitem[{Williams et~al.(2019)Williams, Yazdani, Suresh, Scheutz, and Beetz}]{williams}
Tom Williams, Fereshta Yazdani, Prasanth Suresh, Matthias Scheutz, and Michael Beetz. 2019.
\newblock \href {https://doi.org/10.1007/s10514-018-9795-5} {Dempster-shafer theoretic resolution of referential ambiguity}.
\newblock \emph{Auton. Robots}, 43(2):389–414.

\bibitem[{Wu et~al.(2025)Wu, Wang, Yu, Zhang, Chang, and Yu}]{Wu2024LongMemEvalBC}
Di~Wu, Hongwei Wang, Wenhao Yu, Yuwei Zhang, Kai-Wei Chang, and Dong Yu. 2025.
\newblock \href {https://api.semanticscholar.org/CorpusID:273345961} {Longmemeval: Benchmarking chat assistants on long-term interactive memory}.
\newblock \emph{ICLR}, abs/2410.10813.

\bibitem[{Wu and Yu(2024)}]{wu-yu-2024-stateful}
Qingyang Wu and Zhou Yu. 2024.
\newblock \href {https://aclanthology.org/2024.findings-eacl.57/} {Stateful memory-augmented transformers for efficient dialogue modeling}.
\newblock In \emph{Findings of the Association for Computational Linguistics: EACL 2024}, pages 853--867, St. Julian{'}s, Malta. Association for Computational Linguistics.

\bibitem[{Yang et~al.(2024)Yang, Yang, Zhang, Hui, Zheng, Yu, Li, Liu, Huang, Wei, Lin, Yang, Tu, Zhang, Yang, Yang, Zhou, Lin, Dang, Lu, Bao, Yang, Yu, Li, Xue, Zhang, Zhu, Men, Lin, Li, Tang, Xia, Ren, Ren, Fan, Su, Zhang, Wan, Liu, Cui, Zhang, and Qiu}]{qwen2.5}
An~Yang, Baosong Yang, Beichen Zhang, Binyuan Hui, Bo~Zheng, Bowen Yu, Chengyuan Li, Dayiheng Liu, Fei Huang, Haoran Wei, Huan Lin, Jian Yang, Jianhong Tu, Jianwei Zhang, Jianxin Yang, Jiaxi Yang, Jingren Zhou, Junyang Lin, Kai Dang, and 23 others. 2024.
\newblock Qwen2.5 technical report.
\newblock \emph{arXiv preprint arXiv:2412.15115}.

\bibitem[{Zhang et~al.(2025)Zhang, Dai, Bo, Ma, Li, Chen, Zhu, Dong, and Wen}]{survey_memory}
Zeyu Zhang, Quanyu Dai, Xiaohe Bo, Chen Ma, Rui Li, Xu~Chen, Jieming Zhu, Zhenhua Dong, and Ji-Rong Wen. 2025.
\newblock \href {https://doi.org/10.1145/3748302} {A survey on the memory mechanism of large language model-based agents}.
\newblock \emph{ACM Trans. Inf. Syst.}, 43(6).

\bibitem[{Zhang et~al.(2024)Zhang, Rossi, Kveton, Shao, Yang, Zamani, Dernoncourt, Barrow, Yu, Kim, Zhang, Gu, Derr, Chen, Wu, Chen, Wang, Mitra, Lipka, Ahmed, and Wang}]{Zhang2024PersonalizationOL}
Zhehao Zhang, Ryan~A. Rossi, Branislav Kveton, Yijia Shao, Diyi Yang, Hamed Zamani, Franck Dernoncourt, Joe Barrow, Tong Yu, Sungchul Kim, Ruiyi Zhang, Jiuxiang Gu, Tyler Derr, Hongjie Chen, Ju-Ying Wu, Xiang Chen, Zichao Wang, Subrata Mitra, Nedim Lipka, and 2 others. 2024.
\newblock \href {https://api.semanticscholar.org/CorpusID:273798244} {Personalization of large language models: A survey}.
\newblock \emph{ArXiv}, abs/2411.00027.

\bibitem[{Zheng et~al.(2023)Zheng, Chiang, Sheng, Zhuang, Wu, Zhuang, Lin, Li, Li, Xing, Zhang, Gonzalez, and Stoica}]{judge}
Lianmin Zheng, Wei-Lin Chiang, Ying Sheng, Siyuan Zhuang, Zhanghao Wu, Yonghao Zhuang, Zi~Lin, Zhuohan Li, Dacheng Li, Eric~P. Xing, Hao Zhang, Joseph~E. Gonzalez, and Ion Stoica. 2023.
\newblock Judging llm-as-a-judge with mt-bench and chatbot arena.
\newblock In \emph{Proceedings of the 37th International Conference on Neural Information Processing Systems}, NIPS '23, Red Hook, NY, USA. Curran Associates Inc.

\end{thebibliography}
